# Tex-ViT: A Generalizable, Robust, Texture-based Dual-Branch Cross-Attention Deepfake Detector

Deepak Dagar[1], Dinesh Kumar Vishwakarma[2,*]
Biometric Research Laboratory, Department of Information Technology, Delhi Technological University, Bawana Road, Delhi-110042, India
deepakdagargate@gmail.com[1], dvishwakarma@gmail.com[2,*]

**Abstract**
Deepfakes, generated using Generative Adversarial Networks (GANs) and Variational Autoencoders (VAEs), pose significant threats by altering media through realistic facial modifications. The traditional Convolutional Neural Networks (CNNs) frequently encounter challenges in generalizing across varied datasets because they lack robust features. Vision Transformers (ViTs) demonstrate the potential in image classification; however, their effectiveness is significantly dependent on large training datasets and substantial computational resources, which constrains their practical use. Moreover, existing detection systems are susceptible to adversarial tactics like blurring, compression, and noise introduction, resulting in diminished generalizability and robustness. To address these limitations, we introduce Tex-ViT, an advanced technique that combines ResNet architecture with a cross-attention Vision Transformer framework, augmented by CNN features. Our model incorporates a texture module operating on specified ResNet features before downsampling, leveraging the observation that counterfeit images exhibit surface characteristics differing from authentic images. Through experiments on multiple FaceForensics++ datasets and various GAN datasets, we demonstrate Tex-ViT's superior generalization capabilities, achieving 98% accuracy in cross-domain assessments. The model effectively recognizes both prevalent and distinctive textural characteristics in modified samples, making it a reliable deepfake detector suitable for diverse post-processing contexts.

## 1 Introduction

Advancements in technology, particularly Generative Adversarial Networks (GANs), enable the creation of highly realistic content capable of readily deceiving the unaided eye. Deepfake represents the most advanced forms of visual and auditory manipulation technology. Deepfake is a system that generates remarkably realistic and convincing content through deep learning techniques. Visual deepfakes are categorized into five types: lip synchronization, attribute manipulation, full-image synthesis, body re-enactment, and facial application [1]. Using deepfake technology has benefited the education and entertainment sectors in numerous ways. Nonetheless, its detrimental facets have jeopardized individual reputations, instigated societal turmoil, and endangered national tranquility [1].

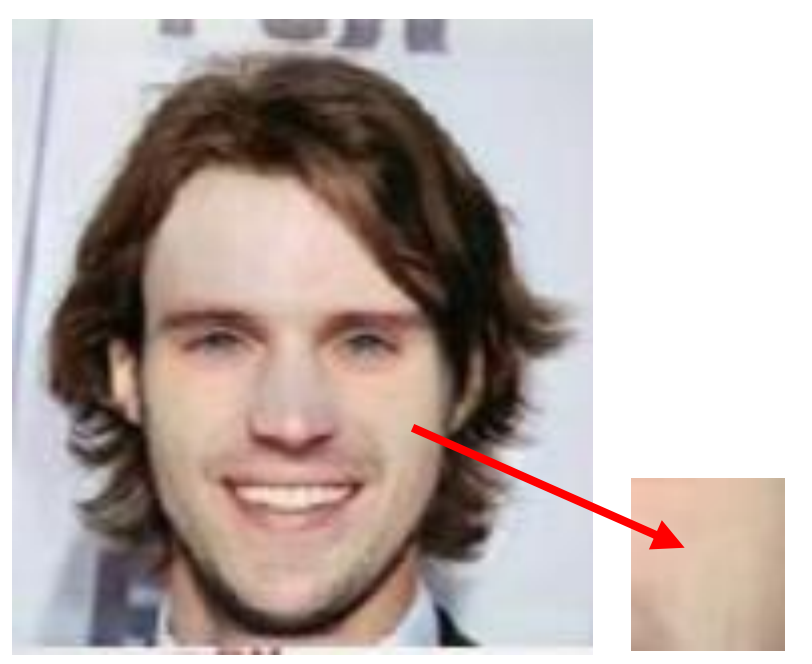
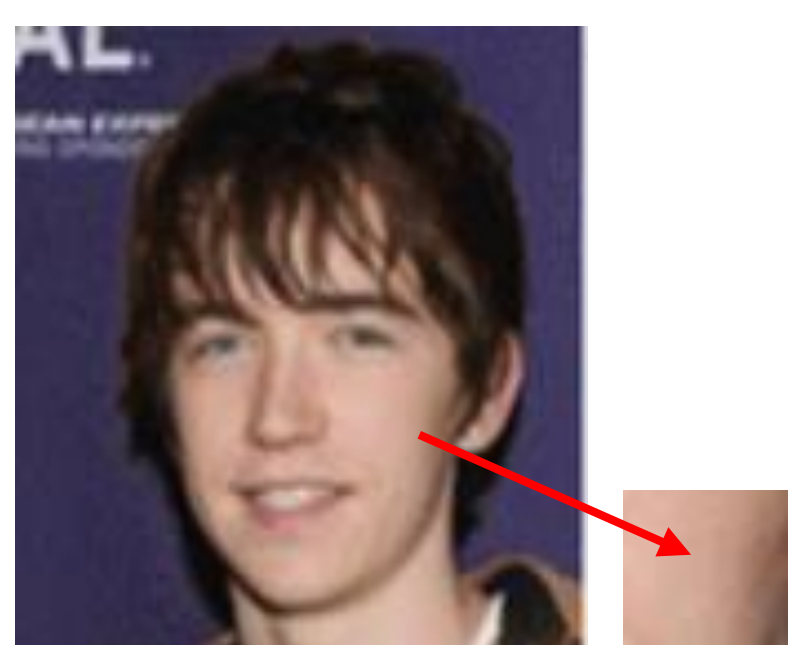
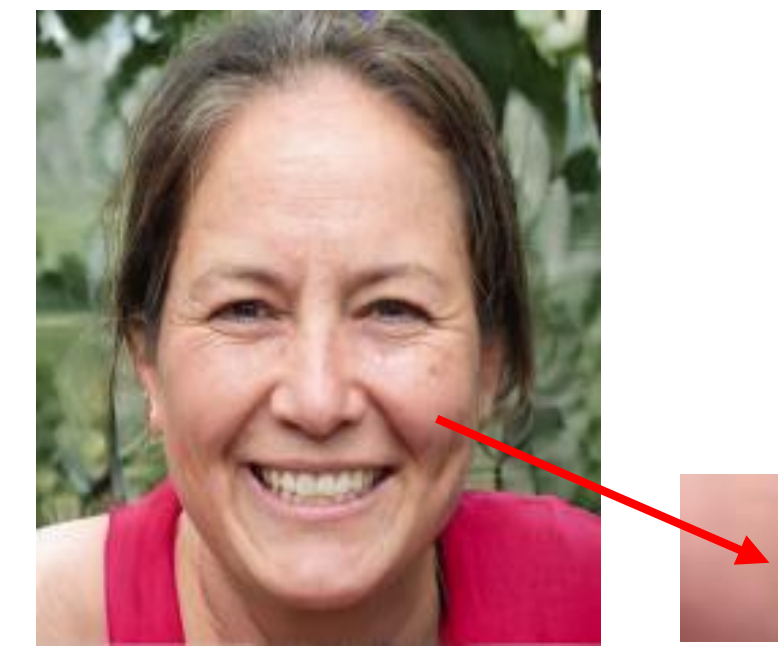

Figure 1: Fake images on a closer look, showing that fake images tend to have smoother surfaces.

The initial research investigations on face forgery detection approached it as a binary classification challenge, focusing on network design to attain effective detection outcomes. Nonetheless, these detectors encounter a significant obstacle in terms of generalization, given the swift advancements in Deepfake technology. The differences in data distribution among various datasets result in suboptimal performance of a detector when applied across datasets, even though it may exhibit high accuracy within a specific dataset. For a model to perform effectively across various distribution datasets, it is essential to identify the shared and consistent characteristics in the manipulated datasets within a cross-domain framework. Texture may serve as a ubiquitous characteristic that persists throughout diverse forms of manipulation. The proposed approach is predicated on counterfeit images exhibiting smoother surfaces and a more consistent overall texture than authentic images(Figure 1) [2]. This happens because generative models struggle to replicate intricate and varied textures in genuine visual data.

The majority of current detectors utilize Convolutional Neural Networks, particularly those with Xception backbones [3], EfficientNet [4], or ResNet [5], as these architectures have proven to be more effective in detecting Deepfakes. Additional investigations enhance generalizability by concentrating on common indicators of forgery in altered faces, such as the integration of noise information [6] [7], blending artifacts [8] [9], and frequency characteristics [10] [11]. However, CNN algorithms with limited receptive fields may inadequately capture the long-term correlations of features. Long-range feature modeling requires the integration of an additional module within the CNN architecture to capture global texture across diverse semantic levels [12]. Consequently, the study of face forgery employs various techniques utilizing Vision Transformers. ViT-based approaches [13] [14] leverage the self-attention mechanism to capture long-range dependencies among input tokens. Nonetheless, ViT demonstrates limitations in effectively capturing the subtleties of local features in the context of Deepfake detection [15]. It is widely understood that training a Vision Transformer model for deepfake detection necessitates a substantial dataset to attain satisfactory generalization outcomes. Furthermore, these methods lead to a deterioration in performance in practical situations when visual input may be compressed, subjected to noise, and face unforeseen blurring. Nonetheless, specific methodologies have partially mitigated these concerns [16] [17] [18]. Additional efforts are required to close this gap.

To address the limitations of generalization to other domains and robustness to various post-processing operations, this paper proposes a Tex-ViT deepfake detection model that employs a texture-based multi-scale cross-attention transformer. The Tex-ViT incorporates ResNet backbone components with conventional features and a distinct texture module that functions concurrently with segments during each down-sampling phase. Following extraction, the texture-specific features from this module proceed to the dual attention components of the cross-attention vision transformer. The necessity to combine CNNs with vision transformers arose due to CNNs' inability to identify global patterns at extensive distances, a limitation stemming from their inherent constraints of translation and local reference invariance, as well as equivariance. ViT addresses the limitation of locality in CNNs, which effectively captures long-range features. Meanwhile, various data augmentation techniques mitigate the need for extensive data samples in ViT. Additionally, Vision transformers employ attention techniques to accurately discern critical global textural elements essential for deepfake detection. The paper's contribution can be summarized as follows:

- An empirical experimentation and observation is done to indicate that texture data is important for distinguishing manipulated media. The investigation also revealed that the simulated images do not retain longer-range texture correlation.
- The researchers utilized their model to examine FF++ and diverse GAN-generated image collections to demonstrate its adaptability across distribution patterns. The experimental results validate that the proposed model outperforms current methodologies by utilizing texture statistics to detect counterfeit samples effectively.
- Researchers tested the model by evaluating the Celeb-DF, FF++, and DFDCPreview datasets using different post-processing methods, including blurring, noise addition, and data compression. The model outperformed other SOTA approaches in these tests, demonstrating its robustness to various conditions.
- Ablation studies were conducted on each model component to illustrate its usefulness and the efficacy of their combination and to establish that texture with a cross-attention mechanism is a potent characteristic.
- The computational complexity analysis is conducted on various advanced computer vision and deepfake detection models, demonstrating that the proposed model is lightweight, including 43 million parameters, and outperforms previous models across multiple scenarios.

The subsequent portion of the manuscript is organized as follows: Section 2 provides comprehensive findings on techniques for generating and detecting deepfakes within the field. Section 3 examines the empirical study to illustrate the efficacy of texture as a classifying characteristic. Section 4 delineates the proposed model and describes its constituent components. Section 5 delineates the experimental setup, data preprocessing, and augmentation methodologies, as well as the experimentation under cross-domain conditions utilizing FF++ and diverse GAN images for various post-processing procedures. Section 7 discusses the importance of the several components of the model in great detail. Section 8 delineates the visualization outcomes for each dataset image.Section 9 presents the Limitations and future work for the proposed model. Section 10 concludes the discussion.

## 2 Related Work

The significant research contributions for deepfake detection are discussed in this section.

### 2.1 Deepfake Image Synthesis

The progression of GAN technology has made considerable advancements in producing realistic images since its introduction by Goodfellow [16]. Deepfake image synthesis can be categorized into identity swap, attribute manipulation, and complete face synthesis [1]. An identity exchange conveys identification from the source to the destination. Face-swap [17] and Fake-App [18] employ auto-encoders and decoders for identity alteration. Their conventional methodology for FS is based on a dual auto-encoder and decoder structure [1].

The first and most notable example of an identity swap is **FaceSwap**, which was employed in developing the FaceForensics++ dataset. Recent face-swapping approaches employ GANs due to their capacity to produce exceptionally realistic faces. In recent years, face swap technology has advanced significantly due to the refinement of deep learning methodologies. StyleSwap [19], the style-based GAN, utilizes an architecture that efficiently integrates the requisite information from both branches. A layer masking branch is incorporated to improve the blending of information. FSGANv2 [20] utilizes the method of reenacting Delaunay triangulation and barycentric coordinates, subsequently constructing a face-blending network to attain a smoothing result. The model [21] employs a 3D-aware face swapping technique that generates high-quality and visually coherent swapped faces from single-view source and target images. It utilizes the strong geometric and textural features of 3D human faces, mapping 2D faces onto the foundational space of a 3D generative model by distinguishing identification and attribute properties within the latent space. Other models also integrate 3D-aware techniques with a mask diffusion process; Zhao et al. [22] employ a mid-point estimation method that utilizes a two-point segmentation approach, enabling the incorporation of identity restrictions to generate high-quality facial images.

Faceshifter [23] is a technique that produces very realistic images using an encoder and generator in a self-supervised framework. **Attribute Manipulation** is a kind of image manipulation that involves altering facial qualities such as expressions, eyes, age, and moustaches inside an image. These methods often utilize Evolutionary Dynamics (ED) or a combination of ED with Generative Adversarial Networks (GANs), incorporating a conditioned attribute. StarGAN [24] and STGAN [25] exemplify this phenomenon over the years. Yang et al. [14] propose a GAN technique that utilizes dilated convolutions to modify the receptive fields of superficial layers in StyleGAN. This method facilitates the transformation of fixed-size small features at superficial layers into larger ones that can accommodate different resolutions, hence improving their capacity to represent unaligned faces accurately. Facial characteristics typically employ the disentanglement of attributes within the latent space, facilitating manipulation in a targeted manner. Liu et al. [26] modify the foundational code in the 'style space' and introduce 'Style Intervention,' an efficient optimisation-based method, to improve the visual fidelity of manipulation results. Huang et al. [27] provide a novel approach termed AdaTrans, which is an adaptive non-linear latent transformation designed for disentangled and conditional face editing. The method deconstructs the manipulation process into several smaller stages, in which the direction and magnitude are ascertained based on facial characteristics and concealed codes. Contemporary studies sometimes neglect to preserve the domain-invariant elements of images, such as the identity of human faces. Liu et al. [28] present an implicit style function (ISF) to quickly perform image-to-image translation across various modes and domains with pre-trained unconditional generators. The ISF modifies the underlying representation of the fundamental code to guarantee that the resulting image falls within the intended visual spectrum.

**Entire image synthesis**, wherein the complete image is generated utilizing advanced GANs. ProGAN [29] and StyleGAN [30] are prominent examples introduced over the years. The authors have improved the design of StyleGAN2 [31] by integrating the normalization method utilized in the generator. Furthermore, they refined the progressive GAN technique by instituting limits on the generator's mapping, facilitating a more regulated transition from latent code to pictures. In StyleGAN3 [32], the authors identify the aliasing effect by treating all signals in the network as continuous, leading to straightforward, universally applicable architectural improvements that prevent the intrusion of undesirable information into the hierarchical synthesis process. Numerous strategies grapple with balancing the precision and diversity of generated samples. Diffusion-based models, such as ConPreDiff [33], have been utilized for diffusion-based synthesis. This method emphasizes the preservation of semantic relationships among adjacent pixels, leading to the creation of high-quality image synthesis. An alternative method uses Latent Diffusion Models to produce high-resolution images. Podell et al. [34] developed the SDXL, a system that incorporates various innovative conditioning approaches and is taureceptivght on diverse aspect ratios. Furthermore, they offer a refinement model that enhances the visual quality of samples produced by SDXL through a post-hoc image-to-image method. Numerous renowned datasets have been introduced: Faceforensics, Faceforensics++, Celeb-DF, DFDC, and Deeperforensics. These datasets employ facial modification techniques or a combination of multiple facial manipulation methods.

### 2.2 Deepfake detection

Since the emergence of this technology, numerous prominent methodologies have sought distinct indicators for fraud detection; despite breakthroughs in their respective disciplines, these systems do not generalize effectively to unseen samples. Current deepfake detection strategies exhibit limitations when evaluated across diverse datasets, resulting in suboptimal performance when applied to various sources of information. The existing detection methods do not offer adequate resistance to various pre-processing operations, such as blurring, compression, and noise addition, which frequently occur in real-world deepfakes. Various detection methodologies have addressed the issue to some degree. Khalid et al. [35] employ approach-based class detection methodologies with a Variational Encoder, trained on authentic human facial images, while categorizing other classes, such as deepfakes, as anomalies. The method depends on the RMSE function to calculate the reconstruction scores of the images. Their performance is ambiguous for the GAN-generated images; therefore, the model lacks generalizability. Li et al. [36] employ facial X-rays to identify facial forgeries better better. The method presupposes the presence of blending boundaries and disaggregates the input image by amalgamating two or more sources. Nonetheless, the method may be ineffective for wholly synthetic images, and an adversary may create an adversarial example capable of evading the detection process, rendering it non-generalizable to diverse manipulations and insufficiently resistant against varied detection scenarios. Wang et al. [37] introduced an innovative method that involves observing neuronal function on a layer-by-layer basis, with these interaction patterns generating tiny artifacts for detection. The algorithm's performance remains inadequate, as adversaries can create adversarial samples that circumvent the detection process, indicating a lack of robustness. Alternative methods ( [38] [39]) utilize co-occurrence matrices derived from picture pixels to calculate disparities over many spatial and spectral band channels, subsequently employing CNNs for the detection of counterfeit faces. Nevertheless, the information contained in the raw data is compromised due to artisanal characteristics.

The generalization of deepfake detectors has been a significant concern, with numerous studies examining the topic [40] [41] [42] [43] [44] [7]. A generalizable detector may operate across several datasets, allowing it to excel on unfamiliar datasets and those subjected to various adversarial manipulations such as noise, blur, and sample compression. Nadimpalli et al. [45] employ a hybrid approach of reinforcement learning and supervised learning. This method allows for selecting top-k augmentations for the test samples, which are averaged for final classification. The model demonstrates better generalizability; its evaluation is confined to the Celeb-Df and FF++ datasets. The generalization model must identify common characteristics across different types of manipulation. Researchers pursued that direction [46] [47]. Yu et al.

[47] employ a module's independently trained particular forgery feature extractor alongside a U-Net architecture to assess the module's validity. Nonetheless, the strategy presupposes the existence of analogous forgery traces, which may be obscured or camouflaged by the opponent. Atama et al. [48] utilize the picture noise residual domain to investigate the influence of spatial and spatiotemporal variables on enhancing generalization; however, their approach relies on high-frequency information, which is inadequate for data compression contexts. Li et al. [49] developed a code identification method that uses a codebook to represent the spatial distribution of genuine and counterfeit images. The model demonstrates effective performance with diverse cross-dataset and compressed images; however, its efficacy with other and unconventional datasets remains ambiguous. The generalizability of the detector presents a significant difficulty that must be investigated for effective operation in diverse adversarial settings.

***Texture-based detection:*** The The intricacies of convolutional neural networks (CNNs) have garnered considerable focus in recent years.. Numerous authors ( [50] [51] [52] [12] ) have investigated textures, as CNN models exhibit an inherent bias towards textures over forms. Yang et al. [51] introduced an innovative Multi-scale Texture Difference Model that offers a stationary characterization of texture difference by integrating data from pixel gradients and intensities. The model exhibits robust performance across several distortions; however, its efficacy markedly diminishes when confronted with unfamiliar manipulations. Bonomi et al. [53] employ Local Derivative Patterns on Three Orthogonal Planes (LDP-TOP) to investigate the textural dynamics within spatial and temporal dimensions. Although the model exhibits commendable performance throughout multiple modifications of FF++, its efficacy remains dubious concerning various types of GAN pictures and heavily compressed video samples. Yadav et al. [54] propose a distinctive end-to-end architecture for detecting facial manipulation, which extracts discriminative "textural" (T) and "manipulation residuals" (MR) characteristics from two branches in varying proportions. The model's efficacy remains uncertain in unfamiliar settings. Gao et al. [55] examine the commonalities among different forged content regarding texture qualities and artifact presence, subsequently designing an ensemble that enhances the model's generalization capabilities. The method's performance is inadequate for extremely compressed settings. Yang et al. [50] investigate the texture disparity between authentic and counterfeit images in visual saliency, further augmented by the guided filter utilizing a saliency map. The approach necessitates substantial data and must be retrained for novel images. Liu et al. [12] utilize global image texture representations through Gram matrices for effective counterfeit image identification. Their efficacy remains ambiguous concerning other deepfake operations.

***Research gaps and Proposed framework:*** Existing deepfake methods often demonstrate limitations in adapting to novel manipulations. Furthermore, these methods lack the necessary robustness when faced with various detection scenarios, including noise addition, compression, and blurring. Existing deepfake detection methods also utilize convolutional neural networks or vision transformers. CNN algorithms with limited receptive fields [56] and inductive biases may inadequately capture the long-term correlations of features, while Vision transformers need a lot of training data to perform well in cross-domain settings. This work presents a texture-based methodology that combines CNNs and multi-scale vision transformers to address these limitations. ViT addresses the limitation of locality in CNNs, which captures long-range features. Conversely, the requirement for a substantial number of data samples in ViT is mitigated through various data augmentation techniques. The proposed methodology employs texture as an invariant attribute, predicated on the concept that counterfeit images typically exhibit smoother surface features than authentic data samples. The texture is calculated with gram matrices derived from ResNet layers, which are then input into ViT and traditional CNN features. The discriminative approach effectively enhances generalizability and resilience against altered artifacts.

## 3 Empirical Observation

The research community's challenge is identifying the shared discriminative traits among various false images created. Numerous researchers conducted tests in this domain [12] [52]; a significant discovery from these experiments was that texture statistics are essential for various CNN-based detectors, such as ResNet, to effectively capture texture variations for discrimination purposes. This section elucidates the significance of texture as a categorization parameter from several viewpoints.

### 3.1 Texture as a clue

Texture denotes the surface's appearance, defined by its fundamental components' form, size, density, and proportional arrangement. In computer vision, it refers to the recurrent presence of the grey pixel level inside the space [51]. There are three reasons for using facial texture as an important task for deepfake detection.

1) The differences in texture between authentic and artificial faces are notably significant, with natural faces exhibiting a unique and distinguishable distribution of textures. An empirical examination of authentic and counterfeit images is conducted to elucidate the disparities in textural properties. Texturized images are produced for both legitimate and counterfeit images via a texturized generation technique, and Figure 2 illustrates that counterfeit images are devoid of texturized details compared to authentic images. Despite the differences in texture across various regions of the face, a general uniformity is present, indicating a geometric coherence or elevated smoothness in the spatial variation, which plays a significant role in facial recognition [57]. Nonetheless, deepfake generation involves altering the source face, or its components, to resemble those of a target individual, thereby compromising the textural integrity of the original face. The discrepancies are particularly evident at the peripheries of the reenactments, especially around the edges of the face-swapped image, as well as in the eyes and mouth during the reenactments. Creating counterfeit altered data samples typically entails pre-processing, face generation, and post-processing operations. Post-processing techniques are typically employed to conceal texture imperfections; counterfeit images often exhibit a more uniform surface and fewer textural features [58]. Consequently, the absence of a texturized surface may serve as an indicator for detecting counterfeit images.
2) The importance of texture features in facial recognition is underscored by the critical role of object texture in object recognition via CNN, as demonstrated in various studies [59]. Gatys et al. [60] suggest that CNNs continue to excel in classifying texture images, even when the overall shape structure is completely compromised. The importance of texture in the context of image restoration and 3D reconstruction is evident, particularly in the realm of face 3D reconstruction [61]. A key objective in this area is to develop realistic facial textural details, which consist of high-frequency components and distinct features that signify identity. Considering the facts regarding face-swapping deepfake technologies, which manipulate faces that may represent different identities, the texture is a significant element that contributes to the distinction between the original images and those that have been cosmetically altered [62].

3) Post-processing methods present challenges in effectively preserving local details, complicating the differentiation between authentic and spoofed data. The prevalent post-processing techniques, including Poisson blending and color constraints, primarily emphasize general facial features rather than delving into specific local details. This focus creates a realistic allure of the image on a broader scale, yet it neglects the essential aspect of localism that ensures coherence at that level. Consequently, Zhao et al. propose that the method for deepfake detection be viewed as a classification task that necessitates a detailed solution [63].

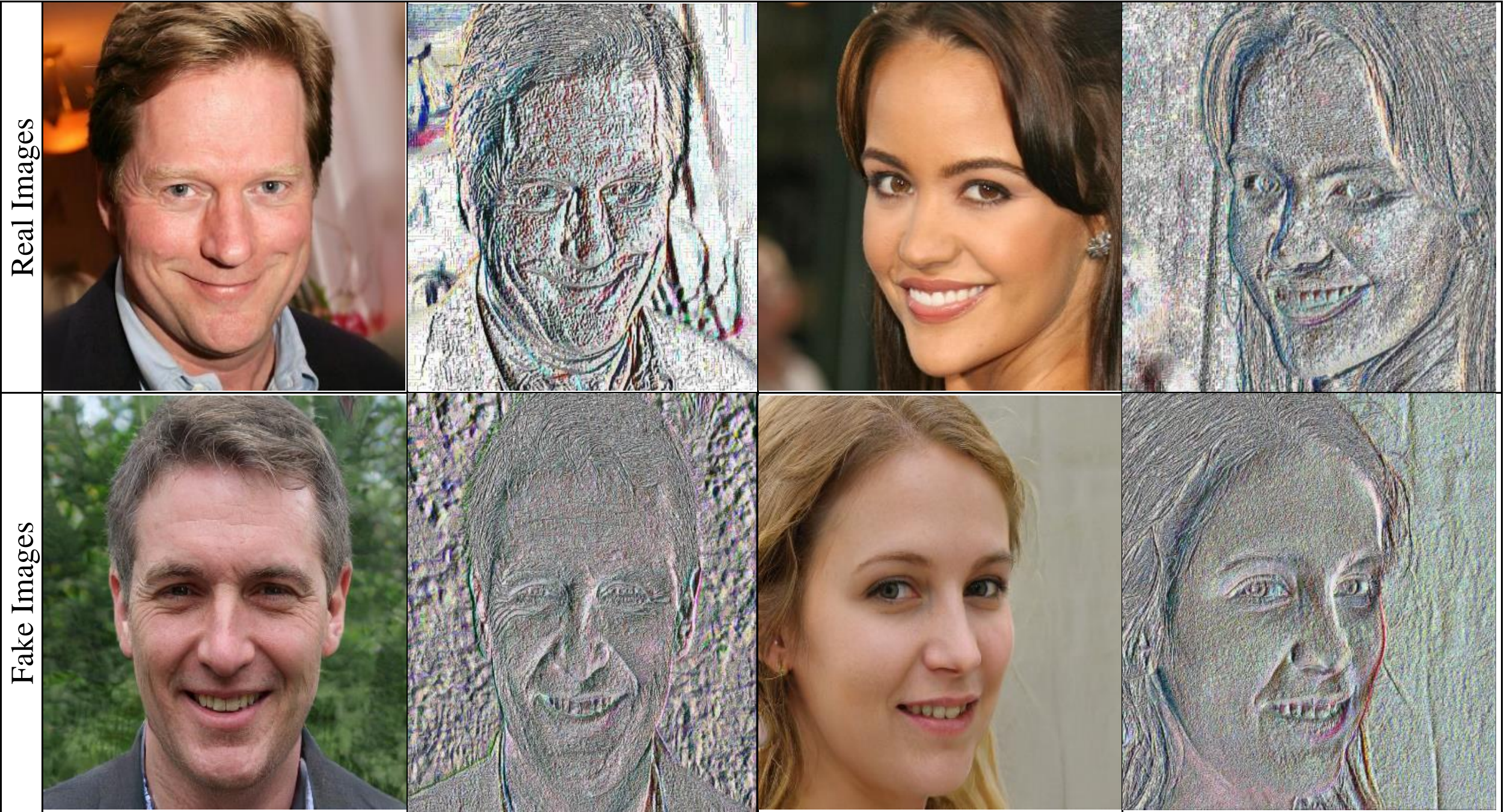

Figure 2: Real and fake Images are shown with their texturized images. Texturized images are generated from the images using the texture-based algorithm.

## 4 Proposed Model

The model consists of two components (Figure 4): texture architecture and a dual-branch cross-attention vision transformer. Texture architecture comprises two branches that function as inputs to the parallel branches of the cross-attention vision transformer.

### 4.1 Texture Architecture:

Texture architecture employs ResNet-18 as its backbone, with the texture block calculated at the input and prior to each down-sampling process, integrating global texture at several layers. The texture block comprises convolutional layers and Gram matrices. Gram matrices are employed to extract texture correlation, involving convolution layers to augment representation. The global texture module is calculated at various semantic levels before each ResNet down-sampling operation to represent long-range texture features [12]. The primary structure of ResNet-18 acquires traditional feature representations of input images at multiple levels using a skip link. It enhances the gradient flow among the layers of multi-scale features, while the texture block acquires the semantic information of global textures across different scales.

*Gram matrices as Texture features:* Gram matrices play an essential role in developing advanced deepfake detectors by leveraging high-level, texture-based feature correlations [64] that remain robust across diverse manipulation conditions and datasets. In contrast to semantic features that heavily rely on facial identity or specific spatial positioning, Gram matrices focus on the relationships among feature channels, serving as a method to represent the second-order statistics of the image. This allows the detector to identify subtle differences in textures, such as unnatural skin smoothness, lighting anomalies, or blending techniques present across various GAN-based fakes, regardless of the specific model or the degree of similarity in their application. The Gram matrices represent the correlations, defined up to a constant factor. Gram matrices $G^{m} \epsilon\ R^{N_m \times N_m}$ evaluate linear dependency among the layers:

$$G_{pq}^{m} = \sum_{r} F_{pr}^{m} F_{qr}^{m} \quad \text{(i)}$$

The Gram matrix is represented by Equation (i). $G_{pq}^{m}$ Denotes the multiplication between the p$^{th}$ feature and q$^{th}$ feature of layer m. In this context, $F^{m}$ indicates the vectorized form of the m$^{th}$ feature map, while $F_{pr}^{m}$ Signifies the k$^{th}$ activation of the p$^{th}$ filter at point r within layer m. Gram matrices integrate feature correlations across spatial dimensions, making them pose-invariant, background-invariant, and occlusion-invariant. This characteristic renders them especially suitable for real-world applications where such variations are inevitable. Furthermore, the decoupling of style (texture) from content (facial structure) allows Gram-based features to reduce overfitting associated with specific individuals or datasets, thereby enhancing the generalization capabilities of the detector [65]. This is particularly beneficial when unrecognized deepfake techniques or novel data distributions must be addressed. Recent studies, such as MixStyle [66] and Gram-Net [12], demonstrate that employing Gram-based features significantly enhances robustness and generalization across different domains. In essence, Gram matrices provide a resilient feature representation that enables the effectiveness of deepfake detectors across multiple manipulations and diverse conditions.

In the proposed model, the texture is produced through the computation of Gram matrices, which the model determines before each down-sampling operation within the ResNet architecture. The matrices are concatenated and subsequently utilized as input for the cross-attention mechanism within the vision transformer.

### 4.2 Dual-branch Cross-Attention Vision Transformer

The vision transformer, devoid of inductive biases, is recognized for its ability to capture long-range, global relationships among pixels, facilitated by its self-attention mechanism and capacity to retain semantic information. The cross-attention module facilitates the exchange of information between these branches by allowing tokens from one scale to attend to tokens from the other. This enables the fusion of complementary information across scales, leveraging the strengths of fine-grained local details and coarse-grained global context. Consequently, the model will develop a more comprehensive image representation, enhancing its generalizability across various object sizes and shapes, ensuring scale invariance, and reducing the likelihood of overfitting to specific visual patterns. The model demonstrates both effectiveness and efficiency, as cross-attention facilitates token-level interactions in a manner that avoids the high computational costs associated with full self-attention. The proposed architecture incorporates two parallel branches, with input patches comparable in scale for each branch. The model processes the input from the texture architecture, subsequently incorporating positional embedding into each patch, including the CLS token, to integrate positional information into the model. Subsequently, the tokens are processed by the stacked transformer encoder. Each transformer encoder features a dual-branch architecture, consisting of Multi-headed Self-Attention (MSA) followed by a Feed-Forward Network (FFN) [67]. FFN consists of two layers within a multi-layer perceptron architecture, employing the GELU activation function for non-linearity subsequent to the initial layer. Layer normalization (LN) is executed subsequent to each block, accompanied by a residual skip connection after every block. The transformer encoder's $\mathrm{x}_0$ ViT and l[th] processing input is represented as follows:

$$\mathrm{x}_0 = [\mathrm{x}_{\mathrm{class}} || \mathrm{x}_{\mathrm{patch}}\mathrm{E}] + \mathrm{E}_{\mathrm{pos}} \qquad E \in R^{P^2.C\,x\,D}, E_{pos} \in R^{(N+1)\,x\,D} \tag{ii}$$

$$z_l = z_{l-1} + MSA\big(LN(\mathrm{x}_{l-1})\big), \qquad l = 1.\ldots..L \tag{iii}$$

$$\mathrm{x}_l = z_l + FFN\big(LN(z_l)\big), \qquad l = 1 \ldots\ldots L \tag{iv}$$

Where class, patch, and positional embedding tokens are represented by $\mathrm{x}_{clsemb} \in \mathrm{R}^{1\times\mathrm{C}}$, $\mathrm{x}_{patchemb} \in \mathrm{R}^{\mathrm{N}\times\mathrm{C}}$ and $\mathrm{x}_{posemb} \in \mathrm{R}^{(\mathrm{N}+1)\times\mathrm{C}}$ respectively. The dimension of the embeddings is represented by C, and N indicates the number of tokens. Consequently, the CLS token from one branch, which has acquired the abstract information, functions as a token query to engage with the patch tokens of the other branch via an attentiveness mechanism, resulting in the production of multi-scale features. The CLS token engages with the patch tokens from the alternate branch. The cross-attention mechanism is illustrated in the following equations, where $\mathrm{x}$ denotes the input to the

$$\mathrm{x}^1 = [\mathrm{x}^1_{cls} || \mathrm{x}^2_{patch}] \qquad \mathrm{x}^1 \in token\ I^{st} branch, \qquad \mathrm{x}^2 \in tokens\ II^{nd}\ branch \tag{v}$$

$$q = \mathrm{x}^1_{cls} W_q, \qquad [k, v] = \mathrm{x}^1 W_{kv} \qquad W_q, W_{kv} \in \mathrm{R}^{\mathrm{D\,x\,3D_h}} \tag{vi}$$

$$A = softmax\left(\frac{qk^T}{\sqrt{D_h}}\right) \qquad A \in R^{NxN} \tag{vii}$$

$$SA(\mathrm{x}^1) = Av \tag{viii}$$

$$MSA(\mathrm{x}^1) = [SA_1(\mathrm{x}^1); SA_2(\mathrm{x}^1); \ldots\ldots; SA_k(\mathrm{x}^1)]W_{msa} \qquad W_{msa} \in R^{k.D_h x D} \tag{ix}$$

$$\acute{y}^1_{cls} = \mathrm{x}^1_{cls} + MSA(LN([\mathrm{x}^1_{cls} || \mathrm{x}^2_{patch}])) \tag{x}$$

In this context, q, k, and v, denotes the query, key, and value accordingly. The term n+1 represents the total number of patches, d represents the model dimension, k represents the number of heads, and Dh(d/k) refers to the dimension of each head. Wq, $W_{kv}$, and $W_{msa}$ represent the learnable parameters associated with the query, key, value, and multi-head self-attention (MSA). After the fusion with other branch tokens, the CLS token within the next transformer encoder re-engages with its patch tokens. This approach enriches the feature reprentation by using the information of other branch for it's corresponding patch tokens (Figure 3). Subsequently, these tokens are processed through the Layer Normalization before being input into the Multi-Layer Perceptron (MLP) for parameter learning.

$$\ddot{y}^1 = \acute{y}^1_{cls} + \mathrm{x}^1_{patch} \tag{xi}$$

$$\check{z} = MLP(LN(\ddot{y}^1)) \tag{xii}$$

Finally, these classification tokens are concatenated for the final predictions. The characteristic of cross attentions plays an essential part in the discriminative and transferring features of cross learning, which the network effectively utilized to improve its classification performance. Furthermore, this mechanism ensures that there is no possibility of overlooking essential features, even when faced with intricate image manipulations necessary for accurate identification. Integrating information across various scales allows the model to detect minor inconsistencies and artifacts that may suggest tampering.

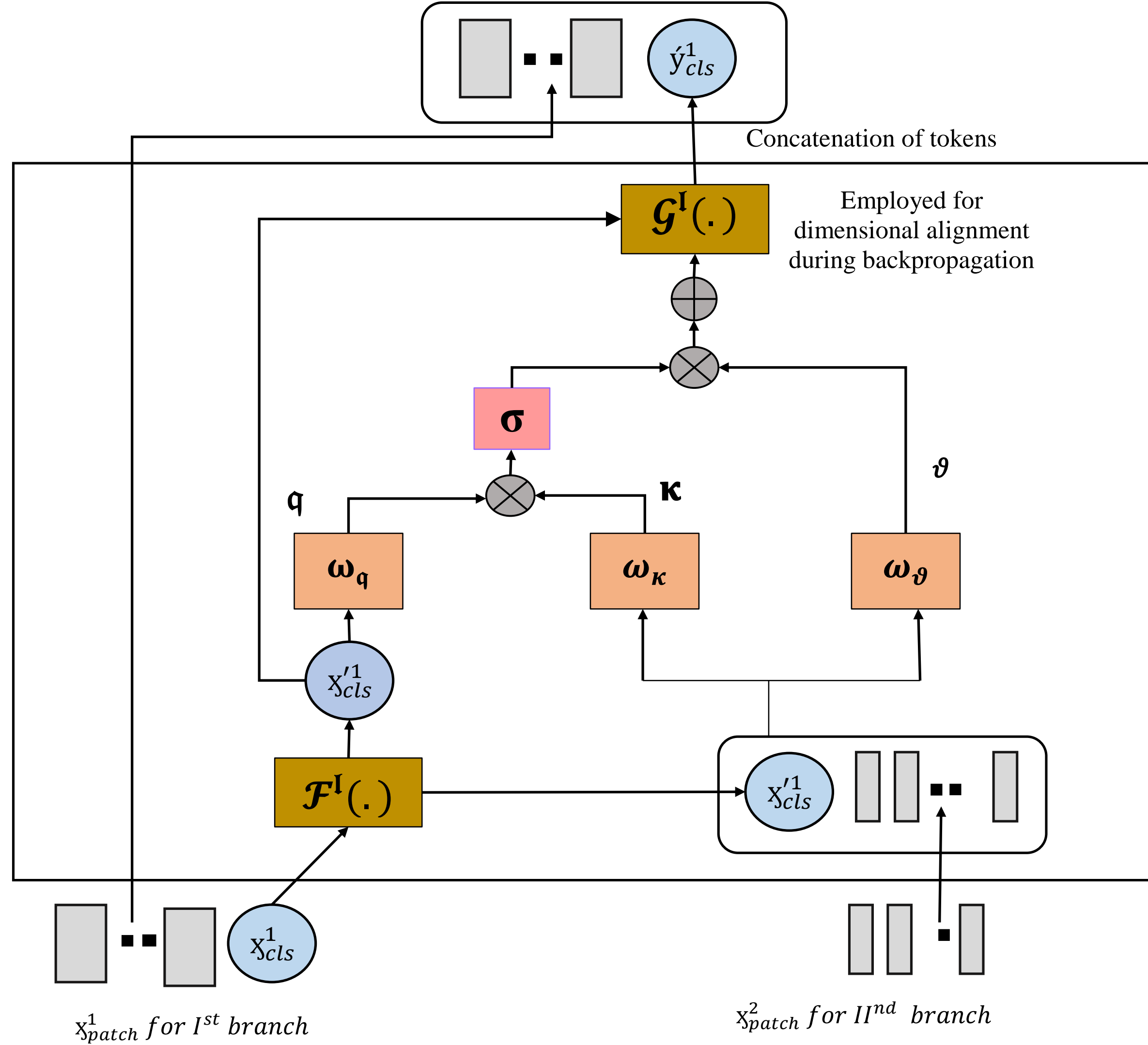

Figure 3: The cross-attention mechanism involves the CLS token from the first branch serving as a query token, facilitating communication with the patch token from the second branch

**Algorithms 1: Tex-ViT for Deepfake classification**

**Parameter Initialisation:**

- Input: Let I = {I1, I2,…..In} represent the collection of images, and L = {0, 1} designates the collection of labels, where 0 corresponds to the real image and 1 corresponds to the deepfake image.
- n represents the size of the dataset.
- Partition I is divided into three subsets: 70% for training, 15% for validation, and 15% for testing.

1: **For 1 to 100 epochs, do**

2: Feed input image I into the ResNet architecture for feature extraction.

3: Calculate the texture utilizing the texture block before each downsampling operation in ResNet, and continue concatenating them.

4: ResNet processed conventional CNNs, and texture features were derived at steps 1 and 2 within the dual branch of the vision transformer.

5: Divide the features into fixed-sized patches and flatten them at each branch.

6: Flatten the image patches to generate linear embeddings in reduced dimensions.

7: Incorporate positional embeddings alongside the CLS token.

8: Input the sequence into the transformer encoder at each branch.
9: Generate tokens by querying the CLS token of the Ist branch alongside patch tokens from an alternate branch, and conversely.
10: Combine the tokens from both branches for classification.
11: Execute the model training in an end-to-end manner and adjust the weights utilizing the Adam optimizer.
12: Assess the validation set and store the weights of the model that demonstrates optimal performance.
13: **end for**
14: Retrieve the weights of the model that were saved at step 11.
15: Assess the performance metrics on the test dataset.

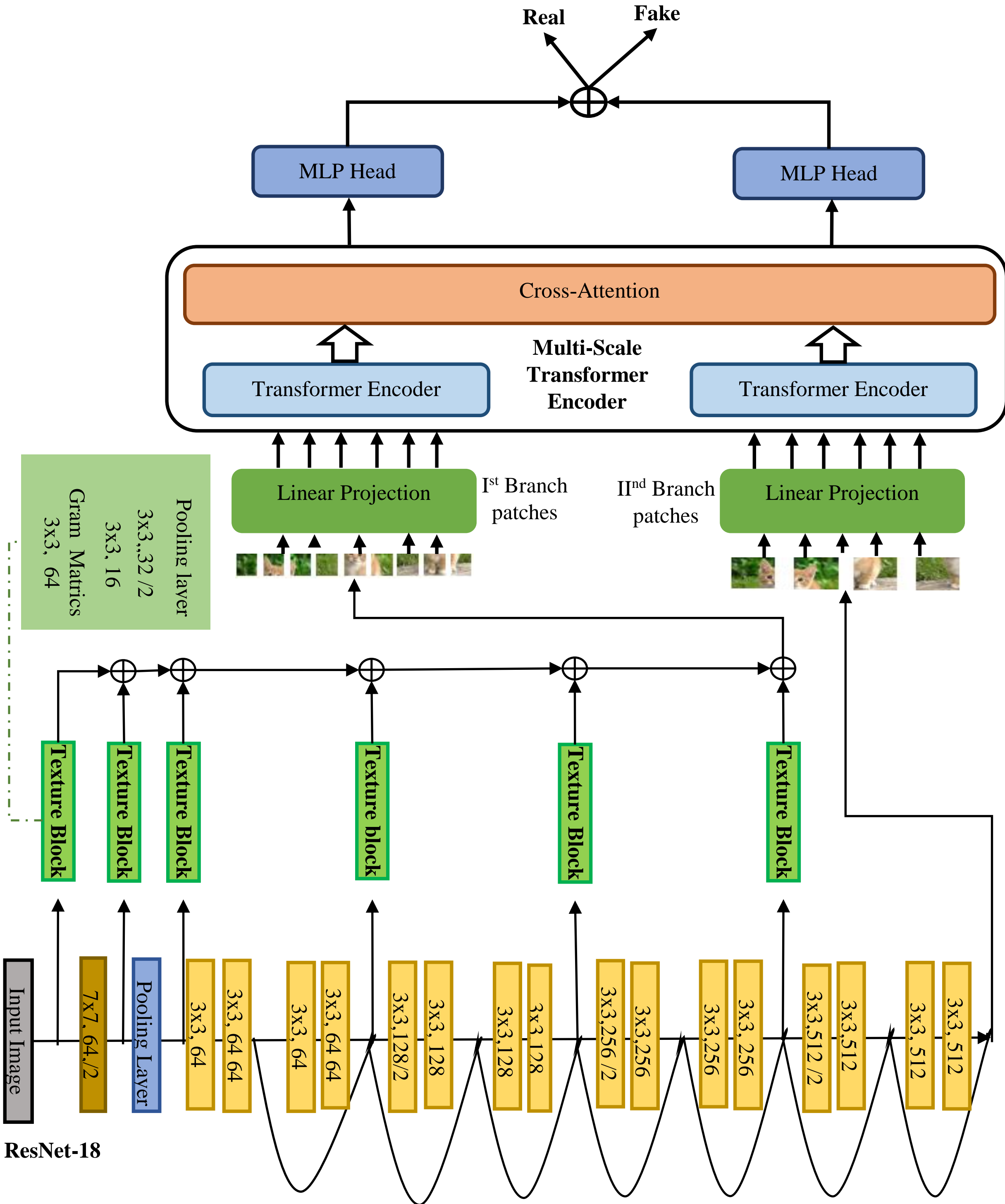

Figure 4: A dual-branch vision transformer processes input from the texture module and the ResNet network.

## 5 Experimentation

This section offers a detailed summary of the selection of training hyperparameters, the datasets employed, the reasoning for choosing the face extractor, and the specific experimental scenarios conducted.

### 5.1 Experimental Settings

The initial learning rate is set at 0.01. In this case, the Adam optimizer is utilized to update the model's parameters due to its extensive usage and efficiency in reducing the time required for parameter updates. Due to the memory constraints, the batch size is configured to 64. The model's performance reaches a saturation point after the specified number of epochs, which is why each experiment is conducted over 100 epochs. The investigations employ 24GB NVIDIA TITAN RTX GPUs.

### 5.2 Dataset Pre-Processing

Three deepfake datasets are utilized for model evaluation: Celeb-DF, DFDC-Preview, and Faceforensics++. The state-of-the-art in current scenarios is represented by these datasets, which are widely used. Frames are extracted from the datasets using the RetinaFaceResNet50 face extractor, which has a lower failure rate than MTCNN, because of its short facial recordings. From each video, a total of one hundred frames are extracted. The DFDC and Celeb-DF datasets predominantly feature faces with aspect ratios ranging from 1 to 1.5, and heights varying between 151 and 200 pixels. Finally, an FF++ has dimensions of [151, 200] and an aspect ratio of [1, 1.5] (refer to Table 1). Various GAN-generated images are utilized for model evaluation. Images generated by ProGAN and StyleGAN and actual image datasets like CelebA-HQ, CelebA, and FFHQ have been retrieved from their respective repositories. The generation of images using StarGAN and STGAN is accomplished by executing the respective code available in their GitHub repositories.

Table 1: Description for the various datasets along with their respective resolutions

| **Deepfake Datasets** | **# of Train. Set Image frames** | **# of Validat. set Image farmes** | **# of Test. set Image frames** | **Image Res.** |
|---|---|---|---|---|
| FF++(DF) | 8000 authtentic, 8000 manipulated image frames. | 2000 authtentic, 2000 manipulated image frames. | 2000 authtentic, 2000 manipulated image frames. | 128x128 |
| FF++(face2face) | 8000 authtentic, 8000 manipulated image frames. | 2000 authtentic, 2000 manipulated image frames. | 2000 authtentic, 2000 manipulated image frames. | 128x128 |
| FF++(Faceswap) | 8000 authtentic, 8000 manipulated image frames. | 2000 authtentic, 2000 manipulated image frames. | 2000 authtentic, 2000 manipulated image frames. | 128x128 |
| FF++(Neural Texture) | 8000 authtentic, 8000 manipulated image frames. | 2000 authtentic, 2000 manipulated image frames. | 2000 authtentic, 2000 manipulated image frames. | 128x128 |
| DFDCPreview | 10,000 authtentic, 8000 manipulated image frames. | 1500 real, 1500 fake image frames. | 1500 real, 1500 fake image frames. | 128x128 |
| Celeb-DF | 10,000 authtentic, 10,000 manipulated image frames. | 8000 authtentic, 8000 manipulated image frames. | 8000 authtentic, 8000 manipulated image frames. | 128x128 |
| ProGAN & CelebA-HQ | 10,000(CelebA-HQ) & 10,000(ProGAN) | 1500(CelebA-HQ) & 1500(ProGAN) | 1500(CelebA-HQ) & 1500(ProGAN) | 1024x1024 |
| StyleGAN & CelebA-HQ | 10,000(CelebA-HQ) & 10,000(StyleGAN) | 1500(CelebA-HQ) & 1500(StyleGAN) | 1500(CelebA-HQ) & 1500(StyleGAN) | 1024x1024 |
| FFHQ and StyleGAN | 10,000(FFHQ) & 10,000(StyleGAN) | 1500(FFHQ) & 1500(StyleGAN) | 1500(FFHQ) & 1500(StyleGAN) | 1024x1024 |
| CelebA & StarGAN | 10,000(CelebA) & 10,000(StarGAN) | 1500(CelebA) & 1500(StarGAN) | 1500(CelebA) & 1500(StarGAN) | 128x128 |
| CelebA & STGAN | 10,000(CelebA) & 10,000(StarGAN) | 1500(CelebA) & 1500(StarGAN) | 1500(CelebA) & 1500(StarGAN) | 128x128 |

### 5.3 Data Augmentation and Scaling

The original ViT model [68] indicates that a vision transformer typically requires training data to achieve performance comparable to a CNN model. DeiT [69] has demonstrated promising results by utilizing a comprehensive array of data-augmentation techniques, achieving comparable performance to the CNN model while requiring fewer data inputs. The proposed model employs several data-augmentation techniques, including rand augmentation [70], cut mix [71], and mixup [72]. Random-erasing [69] and drop path regularization techniques also enhance the overall classification results.

### 5.4 Experiments on the cross-domain settings for the Faceforensics++ dataset

Various categories of Faceforensics++ have undergone experimentation [73] . A single class of FF++ is used to train the model, and it is then tested on both that class and additional FF++ variations. The models' weights that demonstrate high performance on the validation set are preserved for assessment on the test dataset. Four models are considered for the comparison. For the comparison, seven models are taken into consideration:

a) MesoInception-4 model [74].
b) Capsule Network [75].
c) Combining Vision Transformers and Efficient Net(E-ViT) [76].
d) UCF [46].
e) IID [27].
f) SIA[21]
g) UIA[22]

The code for these models has been sourced from their GitHub repository and modified to align with the dataset. Additionally, further evaluation metrics have been incorporated to facilitate a thorough evaluation. After training on a particular manipulation category, the models were tested on several FF++ manipulation categories. The experiments are essential for evaluating the models' performance under different manipulations, which is important for validating the detector's generalization capabilities.

The performance metrics of various models trained on the DeepFakes category of the FF++ dataset, followed by evaluations across the other categories within the same dataset as shown in **Table 2**. Multiple models demonstrate high performance for the same type of manipulation, with some achieving perfect scores, indicating a potential overfitting issue. The performance of these overfitted models significantly deteriorated when tasked with classifying additional categories of FF++. MesoNet and CapsuleNet are recognized as leading models for forgery detection, achieving an accuracy score of around 50% in identifying manipulated images within the FF++ dataset. Their limited capacity to learn only traditional features from convolutional neural networks (CNN) renders them inadequate for effectively detecting cross-manipulation scenarios. The Uia and Ucf approaches, recognized for identifying common traits across various manipulation types, fail to satisfy the requirements for a generalized deepfake detection system. The model demonstrates superior performance compared to alternative models, achieving an accuracy rate of 72% specifically within the DF category of the dataset. Most models exhibit difficulties in accurately classifying the face-swap category within FF++, with a limited number of models achieving performance levels below 50% accuracy. The performance of our model indicates that texture is a reliable attribute that persists across various forms of facial modifications.

The performance metrics of different models trained on the FaceSwap category of the FF++ dataset, together with evaluations across other categories within the same dataset, are presented in **Table 3**. The various models are highly effective within the face2face category; however, their performance markedly decreases when utilized in other manipulation categories. The CapsuleNet approach demonstrates the lowest performance compared to all other methods in cross-manipulation scenarios. In the face2face category, all models demonstrate a marked performance improvement compared to earlier scenarios within cross-manipulation settings. This is because face2face employs facial re-enactment techniques that utilize video frames to produce a detailed reconstruction of the face, considering various lighting conditions and facial expressions. The inherent characteristics of the models facilitate the learning of implicit features related to the manipulation, thereby assisting in cross-manipulation scenarios. The Uia model demonstrates high effectiveness, achieving an accuracy score of approximately 70%. This model employs an unsupervised approach and integrates an inconsistency-aware module to identify discrepancies within the patch-level data. The Sia approach exhibits suboptimal performance as it heavily depends on the attention mechanism, which may occasionally overlook minor abnormalities in the manipulations. Additionally, multiple models demonstrated inadequate performance, with no significant enhancements observed. In contrast, Tex-ViT preserves the overall characteristics, mitigates overfitting when the training and testing datasets originate from the same distribution, and demonstrates effective generalization to different distribution categories. The manipulation resulted in scores of 73% for the DF category and 71% for the NT category, respectively. The FS category score for the manipulation has increased relative to the score in the previous table.

**Table 4** illustrates the model's performance when trained within the FS category and subsequently evaluated on various iterations of FF++. Each model displays overfitting for the specified category and shows insufficient performance across the remaining manipulation categories. This is mostly due to the Faceswap construction approach, which uses facial landmark points to generate a 3D template. The procedure, which requires exact shape mixing and color correction, presents a significant challenge for the detector to identify accurately. The UCF and Uia model demonstrates suboptimal performance, achieving an accuracy score of roughly 50% when subjected to training on the face swap manipulation. This underscores the vulnerability of the model in varying conditions. Sia exhibits suboptimal performance; however, IID shows a marginal performance improvement relative to earlier models. The observed enhancement can be ascribed to the model's ability to acquire implicit and explicit features. Nevertheless, the overall performance remains significantly below the target score. MesoNet and CapsuleNet show inadequate performance; in contrast, the transformer-based model E-ViT displays markedly enhanced performance due to its hybrid architecture, which integrates CNN and ViT to capture long-range dependencies effectively. The model demonstrates challenges in precisely identifying this manipulation category, achieving an accuracy score between 62% and 67%. The face swap-generating mechanism utilizes complex generation techniques that present detection challenges for our model.

**Table 5** illustrates the final category of manipulation, in which the model undergoes training with the NT category of manipulation and is then assessed on several other categories of FF++. The models developed within this particular category of manipulation demonstrate a notable performance relative to other categories of manipulation. The NT generation mechanism employs a texturing technique to analyze the intrinsic properties of the data sample, allowing it to perform effectively in manipulation categories. All models demonstrate similar performance when trained on the NT category of manipulation, suggesting that texture is one of the invariant properties that enhances their capability to operate across various types of manipulation. Most models attained a score exceeding 70% during testing on the DF manipulation category. All models in the face2face category, except for UCF and IID, demonstrate strong performance and even exceed the results of the previous category. The manipulation category (FS) consistently demonstrates suboptimal performance. Additionally, the problem of overfitting remains evident in the NT category. Our model consistently demonstrates superior performance compared to alternative models. The advanced model attains an accuracy score of 77% in face-to-face manipulation and 76% in Deepfake manipulation categories, indicating superior performance relative to alternative models. Other models achieve near-perfect scores when training and testing data originate from the same distribution; however, they demonstrate poor generalization capabilities when applied to different distributions. On the other hand, our model performs well across various manipulations when using the texture module and cross-attention mechanism, thus demonstrating that texture functions as a consistent feature across a range of manipulations.
Nonetheless, each performance is negatively impacted by the FaceSwap category of FF++. Figure 5 presents the ROC curves for several models, showcasing training on F2F images and testing across different categories.

### 5.5 Experiments on the cross-domain settings for the GAN dataset images

Comprehensive evaluations have been conducted on a range of images generated by GANs. Comprehensive image synthesis techniques, including StyleGAN, ProGAN, and attribute manipulation methods from StarGAN and STGAN, have been evaluated for analysis purposes. High-resolution authentic images are sourced from the FFHQ and CelebA-HQ datasets, whereas low-resolution images are obtained from the CelebA dataset. The five image datasets developed for comprehensive and equitable evaluation namely FFHQ StyleGAN CelebA-HQ ProGAN, CelebA-HQ StyleGAN are the high-quality images datasets while CelebA STGAN and CelebA StarGAN are the low-resolution datasets. These datasets include both real and synthetic images. In order to evaluate the outcomes of these datasets, seven advanced models have been selected for comparison:

a) Xception with depth-wise separable convolution(Xception-Net) [3].
b) CNN's generated images are easy to spot now(CNN-Net) [77].
c) Efficient-Net [4].
d) UCF [46].
e) IID [27].
f) SIA[21]
g) UIA[22]

The code has been extracted from the GitHub source and modified to align with the dataset utilized for these models. Furthermore, supplementary evaluation criteria have been integrated to enhance the comparison assessment. The model demonstrates outstanding performance, exceeding the scores obtained by various leading methodologies. Similarly, the performance of FF++ models developed for cross-forgery remains significantly below the optimal score necessary for effective application in real-world scenarios. Table 6 displays the performance metrics of these models across different datasets. The performance metrics of various models demonstrate variability across different datasets; they exhibit good results in one dataset while underperforming in another. Models that have previously been trained on actual images are able to detect them rapidly when the testing dataset contains both false and real images; however, they struggle to recognize the opposite class. The precision value in the first row indicates a model that has been trained on CelebA-HQ ProGAN and evaluated using CelebA-HQ StyleGAN, indicates that most models can correctly identify the CelebA-HQ category. However, there is a challenge in accurately classifying the other category, as evidenced by the recall value. ProGAN and StyleGAN utilize different modification techniques, resulting in varied feature spaces.

As a result, trained models within a specific category demonstrate less than optimal performance when applied to a different category. The CelebA-HQ and FFHQ real datasets exhibit significant similarities in their characteristics, facilitating a model that has been trained on a specific dataset to operate efficiently on an alternative dataset. When integrated with FFHQ StyleGAN, a model that has been trained with CelebAHQ StyleGAN exhibits robust performance. The IID model has demonstrated superior performance across these scenarios. The datasets utilized are entirely synthetic, and the IID model demonstrates a high proficiency in detecting implicit inconsistencies within the feature space. Tex-ViT demonstrates superior performance compared to alternative models across various configurations. The data demonstrates that the model attains the standard distinguishing features, including a consistent overall texture that persists across various modifications, whether generating a complete image or adjusting different attributes. The recall metric score demonstrates that all models struggle to identify ProGAN images accurately. All models consistently assigned a perfect rating to the images generated by StarGAN when evaluated with STGAN, and the reverse was also true. This may clarify the reasons for employing similar counterfeiting methods in their production. Figure 6 illustrates the ROC curves for multiple models, utilizing images trained and tested across various GAN image datasets.

### 5.6 Experiments on the in-domain settings for various post-processing operations of FF++, Celeb-DF, and the DFDCPreview dataset

One limitation of the various models is their insufficient robustness for various post-processing operations, including blurring, compression, noise addition, scaling, and translation. [1]. The primary post-processing operations applied to the test dataset to demonstrate the model's robustness include blurring, compression, and the addition of noise. The Gaussian blur transformation in PyTorch has been applied to blur the images, utilizing a kernel size of 7x7 and a sigma of 25. Parameters have been set to a zero mean and a standard deviation of 0.2 for noise addition. Lastly, image compression has been executed, resulting in quality degradation by a factor of 3 (Figure 7). Models were trained using standard images and evaluated on images subjected to various post-processing operations. Three main deepfake datasets have been selected for assessment. Seven models have been utilized for the comparative assessment.

a) MesoInception-4 model [74].
b) Capsule Network [75].
c) CNN's generated images are easy to spot now(CNN-Net) [77].
d) UCF [46].
e) IID [27].
f) SIA[21]
g) UIA[22]

*The testing dataset does not undergo any post-processing operations.* The initial row of the table displays the outcomes when the image has not been subjected to any processing operations; under these conditions, all models have achieved optimal performance (Table 7). In the absence of post-processing operations, each model within the FF++ dataset exhibits overfitting, resulting in suboptimal performance across a range of post-processing tasks. In comparing FF++ with DFDCPreview and Celeb-DF, the performance of FF++ models is superior. The score for the Celeb-DF dataset is lower, likely due to the presence of high-quality images that closely resemble authentic images. The consistent lighting and texture contribute to an increased level of difficulty in distinguishing between them. MesoNet has received the least performance relative to other competitive approaches, primarily attributable to its inadequate ability to capture the conventional CNN features. The IID approach demonstrates superior performance compared to alternative models, attaining a nearly perfect score for the DFDCPreview dataset. The model's ability to focus on the varying attributes associated with identity renders it especially effective for face-swapping operations.

*Testing dataset undergoes blurring operations:* The second row for each dataset indicates the score associated with blurring operations. The performance for the blurring operation has not significantly degraded, indicating that the blurry images preserve the manipulated artifacts in the non-blurry images. The images have been blurred to a considerable degree. The models' performance in the case of FF++ exhibited a decrease of at least 12% when subjected to blur operations. The performance for the DFDCPreview and Celeb-Df datasets exhibited a minor decrease, approximately 3-6%, with some exceptions noted for specific models. The evaluation of MesoNet using the Celeb-DF and DFDCPreview datasets reveals a notable decline in its effectiveness. The decline in performance can primarily be attributed to its reliance on traditional operations of Convolutional Neural Networks. The performance of the IID model has declined, especially with the FF++ dataset, but the other dataset shows a slight decline in performance. Sia and Uia demonstrate enhanced resilience relative to other models under the influence of blurring operations. Other models experience considerable effects. With a minor performance decline of around 1% to 2% for the DFDCPreview and Celeb-Df datasets, the model has proven to have improved resilience and recovery capabilities.. Furthermore, nearly 12% of the data samples still exhibit discriminatory artifacts despite significant blurring processes applied.

*Testing dataset undergoes compression:* The quality of the images was diminished by a factor of three due to the compression process. Compressing samples for the DFDCPreview and Celeb-DF datasets has a negligible impact on model performance, indicating that the modified artifacts remain largely unaffected by the size reductionThe performance of the other models (including CNN-Net, Sia, IID, and UCF) in the FF++ dataset is notably inadequate, indicating that these models are not the most suitable choices for compression data samples. A variety of manipulations are incorporated into the dataset, which complicates the model's ability to comprehend the extensive distribution of features. The Celeb-DF dataset minimally influenced the efficacy of the models, whereas the DFDCPreview dataset exhibited a somewhat greater effect. Compression generally involves a decrease in the resolution of data sampling, impacting finer details and resulting in distortions, including ringing, banding, blocking, and halo effects. The distortions significantly impair the gradients in the smooth portions. Models focusing on particular artifacts face challenges when dealing with a wide range of complex properties. Models that focus on various artifacts at different levels of detail demonstrate a higher likelihood of accurately classifying complex feature patterns. The model emphasizes complex characteristics, incorporating standard CNN features across various levels and texturing. The system utilizes a cross-attention mechanism to analyze global and local information, leveraging the functionalities of transformers. This allows the model to obtain more detailed and complex features across various levels.

*Addition of Noise to the testing dataset:* The transformation in PyTorch modifies data samples by adding noise, which has a mean of zero and a standard deviation of 0.25. The performance metrics of all the models experienced a notable decline, with the results for IID, MesoNet CNN-Net, Sia, and Uia metrics dropping to as low as 60%. This demonstrates how susceptible these detection techniques are to noise interference. The IID model's performance significantly deteriorates when noise is added to the DFDC and Faceforensics++ datasets. This has led to incorrect classification and has made the model vulnerable to adversarial perturbations. MesoNet exhibits vulnerability to multiple adversarial strategies, evidenced by a notable reduction in its classification score. Noise diminishes image quality by obscuring anomalies or inconsistencies and introducing random variations and patterns. As a result, the process of accurate or complete feature extraction is hindered. A model that does well on a noisy dataset can gain advantages from several sophisticated data augmentation techniques that introduce additional noise, aiding in the model's classification learning. Furthermore, it can utilize an attention mechanism that efficiently harnesses multi-scale features, as this multi-scale approach facilitates a robust representation of various attributes, including both local and global characteristics, which are essential for mitigating the impact of noise. Utilizing the latter approach enables our model to acquire intricate and resilient features capable of withstanding various adversarial operations. The Tex_ViT model consistently outperforms the other models in the DFDCPreview dataset when its accuracy falls below 80%, attaining an accuracy score of 98%. However, the model continues to be vulnerable to introducing disturbances to a certain degree.

Table 2: Models were tested across many categories after being trained on the deepfake dataset of FF++.. Here, bold values represent the highest score among competitive methods.

| Test | Ucf | | | | | IID | | | | | MesoNet | | | | | CapsuleNet | | | | | E-ViT | | | | | Sia | | | | | Uia | | | | | Tex-ViT(ours) | | | | |
|---|---|---|---|---|---|---|---|---|---|---|---|---|---|---|---|---|---|---|---|---|---|---|---|---|---|---|---|---|---|---|---|---|---|---|---|---|---|---|---|---|
| | **P.** | **R.** | **F1** | **A.U.C** | **Accu.** | **P.** | **R.** | **F1** | **A.U.C** | **Accu.** | **P.** | **R.** | **F1** | **A.U.C** | **Accu.** | **P.** | **R.** | **F1** | **A.U.C** | **Accu.** | **P.** | **R.** | **F1** | **A.U.C** | **Accu.** | **P.** | **R.** | **F1** | **A.U.C** | **Accu.** | **P.** | **R.** | **F1** | **A.U.C** | **Accu.** | **P.** | **R.** | **F1** | **A.U.C** | **Accu.** |
| Df | 0.9988 | 0.9986 | 0.9986 | 0.99987 | 0.9988 | **1.0** | **1.0** | **1.0** | **1.0** | **1.0** | 1.0 | 0.999 | 0.9994 | 1.0 | 0.9995 | **1.0** | **1.0** | **1.0** | **1.0** | **1.0** | **1.0** | **1.0** | **1.0** | **1.0** | **1.0** | 0.999 | 1.0 | 0.9995 | 1.0 | 0.9995 | **1.0** | **1.0** | **1.0** | **1.0** | **1.0** | 0.9829 | 0.982 | 0.9824 | 0.9985 | 0.9825 |
| F2F | 0.791 | 0.263 | 0.263 | 0.697 | 0.6023 | 0.7790 | 0.2715 | 0.4026 | 0.7218 | 0.5972 | 0.6667 | 0.1455 | 0.2389 | 0.6345 | 0.5365 | 0.6239 | 0.151 | 0.2431 | 0.6159 | 0.53 | 0.755 | 0.165 | 0.2708 | 0.6479 | 0.5557 | 0.6464 | 0.3455 | 0.4503 | 0.6345 | 0.5783 | 0.7286 | 0.192 | 0.3039 | 0.6508 | 0.5602 | **0.7242** | **0.6126** | **0.6637** | **0.7048** | **0.6948** |
| FS | 0.224 | 0.018 | 0.033 | 0.711 | 0.5396 | 0.4637 | 0.016 | 0.0309 | 0.5621 | 0.4987 | 0.2168 | 0.0155 | 0.0289 | 0.2962 | 0.4797 | 0.3764 | 0.0335 | 0.0615 | 0.4207 | 0.489 | 0.392 | 0.265 | 0.4964 | 0.4700 | 0.4927 | 0.3245 | 0.0865 | 0.1365 | 0.3999 | 0.4532 | 0.4652 | 0.0435 | 0.0709 | 0.5062 | 0.49625 | **0.6952** | **0.040** | **0.07564** | **0.6547** | **0.6291** |
| NT | 0.8164 | 0.238 | 0.368 | 0.6693 | 0.5923 | 0.8765 | 0.245 | 0.3829 | 0.7263 | 0.6025 | 0.7703 | 0.208 | 0.3275 | 0.6972 | 0.573 | 0.7132 | 0.25 | 0.3702 | 0.6639 | 0.5747 | 0.789 | 0.257 | 0.3877 | 0.6829 | 0.5942 | 0.6616 | 0.5025 | 0.5711 | 0.6707 | 0.6227 | 0.7794 | 0.304 | 0.4374 | 0.7097 | 0.609 | **0.7248** | **0.6989** | **0.6541** | **0.7958** | **0.7028** |

Table 3: Models were tested across many categories after being trained on the face2face dataset of FF++.. Here, bold values represent the highest score among competitive methods.

| Test | Ucf | | | | | IID | | | | | MesoNet | | | | | CapsuleNet | | | | | E-ViT | | | | | Sia | | | | | Uia | | | | | Tex-ViT(ours) | | | | |
|---|---|---|---|---|---|---|---|---|---|---|---|---|---|---|---|---|---|---|---|---|---|---|---|---|---|---|---|---|---|---|---|---|---|---|---|---|---|---|---|---|
| | **P.** | **R.** | **F1** | **A.U.C** | **Accu.** | **P.** | **R.** | **F1** | **A.U.C** | **Accu.** | **P.** | **R.** | **F1** | **A.U.C** | **Accu.** | **P.** | **R.** | **F1** | **A.U.C** | **Accu.** | **P.** | **R.** | **F1** | **A.U.C** | **Accu.** | **P.** | **R.** | **F1** | **A.U.C** | **Accu.** | **P.** | **R.** | **F1** | **A.U.C** | **Accu.** | **P.** | **R.** | **F1** | **A.U.C** | **Accu.** |
| Df | 0.8531 | 0.3165 | 0.4617 | 0.7533 | 0.631 | 0.8638 | 0.387 | 0.5345 | 0.8117 | 0.663 | 0.7459 | 0.367 | 0.4919 | 0.7085 | 0.621 | 0.8161 | 0.253 | 0.3862 | 0.6304 | 0.598 | 0.6751 | 0.2775 | 0.3933 | 0.6733 | 0.572 | 0.6217 | 0.544 | 0.5802 | 0.6560 | 0.6065 | 0.7305 | 0.6305 | 0.6768 | 0.777 | 0.6989 | **0.7365** | **0.664** | **0.6983** | **0.7958** | **0.7133** |
| F2F | 0.9975 | 0.9985 | 0.9980 | 0.9986 | 0.998 | 0.9995 | 1.0 | 0.9997 | 1.0 | 0.9997 | **1.0** | **1.0** | **1.0** | **1.0** | **1.0** | **1.0** | **1.0** | **1.0** | **1.0** | **1.0** | **1.0** | **1.0** | **1.0** | **1.0** | **1.0** | 0.9960 | 0.9995 | 0.9977 | 0.9999 | 0.9977 | **1.0** | **1.0** | **1.0** | **1.0** | **1.0** | 0.9714 | 0.9875 | 0.9794 | 0.9986 | 0.9793 |
| FS | 0.644 | 0.124 | 0.2079 | 0.5282 | 0.5277 | 0.5780 | 0.1315 | 0.2143 | 0.6605 | 0.5177 | 0.5871 | 0.266 | 0.3662 | 0.6064 | 0.5395 | 0.6371 | 0.151 | 0.2441 | 0.566 | 0.5325 | 0.6399 | 0.2390 | 0.3480 | 0.5922 | 0.552 | 0.5676 | 0.5015 | 0.5325 | 0.6079 | 0.5597 | 0.7120 | 0.4315 | 0.5373 | 0.6473 | 0.6284 | **0.6952** | **0.5982** | **0.6433** | **0.6987** | **0.6593** |
| NT | 0.8092 | 0.314 | 0.4524 | 0.6827 | 0.62 | 0.7891 | 0.378 | 0.5111 | 0.7877 | 0.6384 | 0.6937 | 0.3375 | 0.4541 | 0.6843 | 0.5942 | 0.8122 | 0.2855 | 0.4224 | 0.6666 | 0.6097 | 0.7209 | 0.4855 | 0.5802 | 0.7106 | 0.649 | 0.6741 | 0.694 | 0.6839 | 0.7217 | 0.6792 | 0.7523 | 0.638 | 0.6904 | 0.7949 | 0.7139 | **0.7604** | **0.689** | **0.7229** | **0.8152** | **0.7360** |

Table 4: Models were tested across many categories after being trained on the Faceswap dataset of FF++.. Here, bold values represent the highest score among competitive methods.

| Test | Ucf | | | | | IID | | | | | MesoNet | | | | | CapsuleNet | | | | | E-ViT | | | | | Sia | | | | | Uia | | | | | Tex-ViT(ours) | | | | |
|---|---|---|---|---|---|---|---|---|---|---|---|---|---|---|---|---|---|---|---|---|---|---|---|---|---|---|---|---|---|---|---|---|---|---|---|---|---|---|---|---|
| | **P.** | **R.** | **F1** | **A.U.C** | **Accu.** | **P.** | **R.** | **F1** | **A.U.C** | **Accu.** | **P.** | **R.** | **F1** | **A.U.C** | **Accu.** | **P.** | **R.** | **F1** | **A.U.C** | **Accu.** | **P.** | **R.** | **F1** | **A.U.C** | **Accu.** | **P.** | **R.** | **F1** | **A.U.C** | **Accu.** | **P.** | **R.** | **F1** | **A.U.C** | **Accu.** | **P.** | **R.** | **F1** | **A.U.C** | **Accu.** |
| Df | 0.371 | 0.0305 | 0.0564 | 0.4577 | 0.4894 | 0.7414 | 0.3885 | 0.5098 | 0.7262 | 0.6226 | 0.5892 | 0.522 | 0.5535 | 0.6321 | 0.579 | 0.5375 | 0.0895 | 0.1534 | 0.5415 | 0.5065 | 0.6872 | 0.311 | 0.4282 | 0.6712 | 0.5847 | 0.6432 | 0.274 | 0.3843 | 0.6093 | 0.5610 | 0.5309 | 0.253 | 0.3427 | 0.5789 | 0.5147 | **0.7746** | **0.2486** | **0.3764** | **0.6847** | **0.6248** |
| F2F | 0.7403 | 0.1625 | 0.2665 | 0.5369 | 0.5527 | **0.6943** | **0.4315** | **0.5322** | **0.6964** | **0.6207** | 0.6216 | 0.5455 | 0.5811 | 0.6354 | 0.6067 | 0.6383 | 0.1985 | 0.3028 | 0.5568 | 0.5430 | 0.6700 | 0.3005 | 0.4149 | 0.573 | 0.5762 | 0.7391 | 0.35 | 0.4751 | 0.6554 | 0.6132 | 0.6722 | 0.4605 | 0.5465 | 0.6516 | 0.618 | 0.7546 | 0.2283 | 0.3505 | 0.6446 | 0.6078 |
| FS | 0.9969 | 0.9949 | 0.9949 | 0.9949 | 0.995 | **1.0** | **1.0** | **1.0** | **1.0** | **1.0** | **1.0** | **1.0** | **1.0** | **1.0** | **1.0** | **1.0** | **1.0** | **1.0** | **1.0** | **1.0** | **1.0** | **1.0** | **1.0** | **1.0** | **1.0** | **1.0** | **1.0** | **1.0** | **1.0** | **1.0** | **1.0** | **1.0** | **1.0** | **1.0** | **1.0** | 0.9853 | 0.9755 | 0.9804 | 0.9982 | 0.9805 |
| NT | 0.6137 | 0.058 | 0.105 | 0.4929 | 0.5107 | 0.6113 | 0.2855 | 0.3892 | 0.5926 | 0.552 | 0.5903 | 0.539 | 0.5635 | 0.6205 | 0.5825 | 0.5342 | 0.1365 | 0.2174 | 0.5350 | 0.5087 | 0.6246 | 0.2355 | 0.3420 | 0.5574 | 0.547 | 0.5770 | 0.219 | 0.3175 | 0.5467 | 0.5292 | 0.5275 | 0.2585 | 0.3469 | 0.5181 | 0.5135 | **0.7589** | **0.6378** | **0.6931** | **0.7088** | **0.6728** |

Table 5: Models were tested across many categories after being trained on the deepfake dataset of FF++. Here, bold values represent the highest score among competitive methods.

| Test | Ucf | | | | | IID | | | | | MesoNet | | | | | CapsuleNet | | | | | E-ViT | | | | | Sia | | | | | Uia | | | | | Tex-ViT(Ours) | | | | |
|---|---|---|---|---|---|---|---|---|---|---|---|---|---|---|---|---|---|---|---|---|---|---|---|---|---|---|---|---|---|---|---|---|---|---|---|---|---|---|---|---|
| | **P.** | **R.** | **F1** | **A.U.C** | **Accu.** | **P.** | **R.** | **F1** | **A.U.C** | **Accu.** | **P.** | **R.** | **F1** | **A.U.C** | **Accu.** | **P.** | **R.** | **F1** | **A.U.C** | **Accu.** | **P.** | **R.** | **F1** | **A.U.C** | **Accu.** | **P.** | **R.** | **F1** | **A.U.C** | **Accu.** | **P.** | **R.** | **F1** | **A.U.C** | **Accu.** | **P.** | **R.** | **F1** | **A.U.C** | **Accu.** |
| Df | 0.8814 | 0.613 | 0.7231 | 0.7860 | 0.7652 | **0.7941** | **0.729** | **0.7601** | **0.8542** | **0.77** | 0.7347 | 0.8145 | 0.7725 | 0.8287 | 0.7602 | 0.7619 | 0.616 | 0.6812 | 0.7801 | 0.7117 | 0.7894 | 0.6505 | 0.7133 | 0.8325 | 0.7385 | 0.7236 | 0.6335 | 0.6755 | 0.7699 | 0.6957 | 0.6966 | 0.6395 | 0.6668 | 0.7383 | 0.6805 | 0.7405 | 0.8045 | 0.7714 | 0.8394 | 0.7612 |
| F2F | 0.7989 | 0.379 | 0.5146 | 0.6384 | 0.642 | 0.7169 | 0.6245 | 0.6675 | 0.7843 | 0.6899 | 0.7276 | 0.8615 | 0.7889 | 0.8342 | 0.7695 | 0.7100 | 0.5045 | 0.5898 | 0.7265 | 0.6492 | 0.7583 | 0.739 | 0.7485 | 0.8374 | 0.7517 | 0.7369 | 0.7565 | 0.7466 | 0.8053 | 0.7432 | 0.7532 | 0.748 | 0.7506 | 0.8183 | 07514 | **0.7672** | **0.7995** | **0.7830** | **0.8558** | **0.7785** |
| FS | 0.2403 | 0.0465 | 0.078 | 0.4005 | 0.4497 | 0.3039 | 0.1295 | 0.1816 | 0.4045 | 0.4156 | 0.5013 | 0.3745 | 0.4287 | 0.5296 | 0.501 | 0.4557 | 0.1675 | 0.2449 | 0.4886 | 0.4837 | 0.4966 | 0.2235 | 0.3082 | 0.5057 | 0.4985 | 0.4798 | 0.304 | 0.3722 | 0.5189 | 0.4872 | 0.5543 | 0.429 | 0.4836 | 0.5606 | 0.542 | **0.7072** | **0.5844** | **0.6399** | **0.8148** | **0.6408** |
| NT | 0.9954 | 0.993 | 0.9945 | 0.9972 | 0.9945 | 0.9494 | 0.9955 | 0.9719 | 0.9999 | 0.9712 | 0.9985 | 1.0 | 0.9992 | 1.0 | 0.9992 | 1.0 | 0.9925 | 0.9959 | 0.9978 | 0.996 | **1.0** | **1.0** | **1.0** | **1.0** | **1.0** | **1.0** | **1.0** | **1.0** | **1.0** | **1.0** | **1.0** | **1.0** | **1.0** | **1.0** | **1.0** | 0.9645 | 0.9795 | 0.9719 | 0.9971 | 0.9718 |

Table 6: Models trained and tested on the GAN datasets. Here, bold values represent the highest score among competitive methods.

| Train | Test | Xception | | | | | Ucf | | | | | IID | | | | | CNN-Net | | | | | Efficient-Net | | | | | Sia | | | | | Uia | | | | | Tex-ViT(ours) | | | | |
|---|---|---|---|---|---|---|---|---|---|---|---|---|---|---|---|---|---|---|---|---|---|---|---|---|---|---|---|---|---|---|---|---|---|---|---|---|---|---|---|---|---|
| | | **P.** | **R.** | **F1** | **A.U.C** | **Accu.** | **P.** | **R.** | **F1** | **A.U.C** | **Accu.** | **P.** | **R.** | **F1** | **A.U.C** | **Accu.** | **P.** | **R.** | **F1** | **A.U.C** | **Accu.** | **P.** | **R.** | **F1** | **A.U.C** | **Accu.** | **P.** | **R.** | **F1** | **A.U.C** | **Accu.** | **P.** | **R.** | **F1** | **A.U.C** | **Accu.** | **P.** | **R.** | **F1** | **A.U.C** | **Accu.** |
| CelebA-HQ_ProGAN | CelebAHQ_StyleGAN | 1.0 | 0.06 | 0.113 | 0.7899 | 0.53 | 0.9 | 0.006 | 0.0119 | .4645 | 0.50266 | 1.0 | 0.0033 | 0.0066 | 0.6842 | 0.50166 | 0.9714 | 0.1913 | 0.3212 | 0.9708 | 0.5957 | 1.0 | 0.008 | 0.0158 | 0.8466 | 0.504 | 1.0 | 0.165 | 0.2837 | 0.9378 | 0.5826 | 0.8 | 0.0053 | 0.0105 | 0.8654 | 0.502 | 0.8807 | **0.64** | **0.7413** | **0.8925** | **0.7767** |
| | FFHQ_StyleGAN | 0.904 | 0.0506 | 0.0959 | 0.7546 | 0.5226 | 0.652 | 0.997 | 0.7884 | 0.7777 | 0.7169 | 0.8 | 0.0026 | 0.0053 | 0.6048 | 0.501 | 0.6098 | 0.3466 | 0.4465 | 0.6533 | 0.5703 | 0.7215 | 0.038 | 0.0721 | 0.7205 | 0.5116 | 0.7785 | 0.1546 | 0.2581 | 0.7254 | 0.5553 | 0.9666 | 0.0193 | 0.0379 | 0.7285 | 0.5093 | **0.7695** | **0.6386** | **0.6980** | **0.8091** | **0.7237** |
| CelebA-HQ_StyleGAN | CelebAHQ_ProGAN | 1.0 | 0.024 | 0.0468 | 0.6872 | 51.2 | 0.471 | 0.005 | 0.0105 | 0.4161 | 0.4966 | 0.5 | 0.5 | 0.5 | 0.5468 | 0.5 | 0.7889 | 0.0033 | 0.0066 | 0.8363 | 0.5013 | 1.0 | 0.006 | 0.0013 | 0.8093 | 0.5033 | 1.0 | 0.0053 | 0.0106 | 0.8450 | 0.5026 | 0.5 | 0.5 | 0.5 | 0.4708 | 0.5 | **0.7134** | **0.5445** | **0.6176** | **0.6818** | **0.6065** |
| | FFHQ_StyleGAN | 0.8625 | 1.0 | 0.9262 | 0.9886 | 0.9203 | 0.732 | 0.999 | 0.8452 | 0.7735 | 0.8169 | 0.8631 | **1.0** | **0.9265** | **0.9998** | **0.9206** | 0.9931 | 0.9986 | 0.8619 | 0.9943 | 0.84 | 0.7420 | 0.9993 | 0.8517 | 0.9972 | 0.8260 | 0.6105 | 1.0 | 0.7581 | 0.9783 | 0.6810 | 0.7022 | 1.0 | 0.8251 | 0.9965 | 0.788 | 0.8409 | 0.948 | 0.8912 | 0.945 | 0.8843 |
| FFHQ_StyleGAN | CelebAHQ_ProGAN | 0.9806 | 0.2366 | 0.3813 | 0.8118 | 0.616 | 0.5 | 0.5 | 0.5 | 0.5 | 0.5 | 0.6666 | 0.0013 | 0.0026 | 0.6027 | 0.5033 | 0.7727 | 0.012 | 0.023 | 0.7799 | 0.506 | 0.9032 | 0.0186 | 0.0365 | 0.7952 | 0.5083 | 0.6363 | 0.0046 | 0.0092 | 0.5773 | 0.501 | 0.9643 | 0.036 | 0.0694 | 0.7862 | 0.51733 | 0.8421 | **0.3626** | **0.506** | **0.7454** | **0.6473** |
| | CelebAHQ_StyleGAN | 0.9946 | 0.9993 | 0.9970 | 0.9998 | 0.9970 | 0.996 | 0.99 | 0.9933 | 0.9979 | 0.9933 | 1.0 | **0.9993** | **0.9996** | **0.9999** | **0.9996** | 0.9998 | 0.9963 | 0.9963 | 0.9999 | 0.9963 | 0.9963 | 0.9993 | 0.9993 | 0.9999 | 0.9993 | 0.9986 | 0.9973 | 0.9979 | 0.9999 | 0.998 | 0.9986 | 0.998 | 0.9983 | 0.9999 | 0.9983 | 0.9819 | 0.9793 | 0.9806 | 0.9968 | 0.9807 |
| CelebA StarGAN | CelebA STGAN | 0.9993 | 1.0 | 0.9996 | 1.0 | 0.9996 | 0.999 | 1.0 | 0.9996 | 1.0 | 0.9996 | 1.0 | 1.0 | 1.0 | 1.0 | 1.0 | 1.0 | 1.0 | 1.0 | 1.0 | 1.0 | 1.0 | 1.0 | 1.0 | 1.0 | 1.0 | 1.0 | 1.0 | 1.0 | 1.0 | 1.0 | 1.0 | 1.0 | 1.0 | 1.0 | 1.0 | 0.9775 | 0.927 | 0.9517 | 0.9882 | 0.9530 |
| CelebA StarGAN | CelebA STGAN | 0.9986 | 1.0 | 0.9993 | 0.9999 | 0.9993 | 1 | 1 | 1 | 1 | 1 | 1.0 | 1.0 | 1.0 | 1.0 | 1.0 | 1.0 | 1.0 | 1.0 | 1.0 | 1.0 | 1.0 | 1.0 | 1.0 | 1.0 | 1.0 | 1.0 | 1.0 | 1.0 | 1.0 | 1.0 | 1.0 | 1.0 | 1.0 | 1.0 | 1.0 | 0.9958 | 0.9493 | 0.9720 | 0.9961 | 0.9727 |

Table 7: Models trained on various datasets and tested under various conditions. Here, bold values represent the highest score among competitive methods.

| Training Dataset | Operation | Ucf | | | | | IID | | | | | MesoNet | | | | | CapsuleNet | | | | |
|---|---|---|---|---|---|---|---|---|---|---|---|---|---|---|---|---|---|---|---|---|---|
| | | P. | R. | F1 | A.U.C | Accu. | P. | R. | F1 | A.U.C | Accu. | P. | R. | F1 | A.U.C | Accu. | P. | R. | F1 | A.U.C | Accu. |
| FF++ | No operation | 0.988 | 0.9915 | 0.9898 | 0.9968 | 0.9897 | 0.9897 | 0.9951 | 0.9923 | 0.9996 | 0.9923 | 0.8322 | 0.9998 | 0.9084 | 0.9968 | 0.8991 | **0.9925** | **0.9993** | **0.9993** | **0.9997** | **0.9993** |
| FF++ | Blur | 0.673 | 0.9282 | 0.7803 | 0.8297 | 0.7386 | 0.575 | 0.9975 | 0.7299 | 0.8699 | 0.6304 | 0.5305 | 1.0 | 0.6933 | 0.9819 | 0.5575 | 0.628 | 0.9975 | 0.7715 | 0.8788 | 0.7039 |
| FF++ | Addition of Noise | 0.642 | 0.6445 | 0.6433 | 0.6479 | 0.6415 | 0.5 | 0.5 | 0.5 | 0.5903 | 0.5 | 0.5235 | 0.9928 | 0.6854 | 0.7079 | 0.5445 | 0.6981 | 0.046 | 0.0092 | 0.5612 | 0.5013 |
| FF++ | Compression | 1.0 | 0.0025 | 0.0044 | 0.5784 | 0.5011 | 1.0 | 0.0094 | 0.0186 | 0.912 | 0.5047 | **0.9262** | **0.9868** | **0.9555** | **0.9940** | **0.9541** | 1.0 | 0.0032 | 0.0064 | 0.6939 | 0.5016 |
| DFDCPreview | No operation | 1.0 | 0.9996 | 0.9997 | 0.9997 | 0.9998 | **1.0** | **1.0** | **1.0** | **1.0** | **1.0** | 0.9900 | 0.9993 | 0.9947 | 0.9999 | 0.9946 | 1.0 | 0.9993 | 0.9996 | 0.9999 | 0.9996 |
| DFDCPreview | Blur | 0.998 | 0.7926 | 0.8838 | 0.9287 | 0.8956 | 1.0 | 0.8493 | 0.9185 | 0.9940 | 0.9246 | 0.9985 | 0.8913 | 0.9418 | 0.9981 | 0.945 | 1.0 | 0.748 | 0.8558 | 0.9881 | 0.874 |
| DFDCPreview | Addition of Noise | 0.5 | 1.0 | 0.6666 | 0.5447 | 0.5 | 0.5 | 1.0 | 0.6666 | 0.50 | 0.5 | 0.5 | 1.0 | 0.6666 | 0.5 | 0.5 | 0.4999 | 0.9986 | 0.6662 | 0.7179 | 0.4996 |
| DFDCPreview | Compression | 0.757 | 0.9999 | 0.8619 | 0.7940 | 0.84 | 0.9375 | 1.0 | 0.9677 | 0.9999 | 0.9666 | 0.9695 | 0.9986 | 0.9839 | 0.9994 | 0.9836 | 0.5468 | 1.0 | 0.7070 | 0.9304 | 0.5856 |
| Celeb-DF | No operation | 0.954 | 0.9543 | 0.9544 | 0.9605 | 0.9544 | 0.9227 | 0.9451 | 0.9338 | 0.9773 | 0.9330 | 0.8222 | 0.9075 | 0.8628 | 0.9259 | 0.8556 | 0.9418 | 0.9462 | 0.9441 | 0.9741 | 0.9439 |
| Celeb-Df | Blur | 0.830 | 0.9511 | 0.8865 | 0.8937 | 0.8782 | 0.7483 | 0.9397 | 0.8332 | 0.9174 | 0.8118 | 0.6291 | 0.9757 | 0.7649 | 0.8822 | 0.7002 | 0.8793 | 0.756 | 0.8130 | 0.8979 | 0.8261 |
| Celeb-Df | Addition of Noise | 0.5 | 1.0 | 0.6666 | 0.5083 | 0.50 | 0.6241 | 0.5 | 0.5 | 0.5427 | 0.50 | 0.5 | 0.5 | 0.5 | 0.5267 | 0.50 | 0.5 | 0.5 | 0.5 | 0.5696 | 0.5 |
| Celeb-Df | Compression | 0.973 | 0.5248 | 0.6819 | 0.9124 | 0.7551 | 0.9048 | 0.7524 | 0.8216 | 0.9321 | 0.8366 | 0.9468 | 0.334 | 0.4938 | 0.9193 | 0.6576 | 0.9278 | 0.5146 | 0.6620 | 0.8951 | 0.7373 |

| Training Dataset | Operation | CNN-Net | | | | | Sia | | | | | Uia | | | | | Tex-ViT(ours) | | | | |
|---|---|---|---|---|---|---|---|---|---|---|---|---|---|---|---|---|---|---|---|---|---|
| | | P. | R. | F1 | A.U.C | Accu. | P. | R. | F1 | A.U.C | Accu. | P. | R. | F1 | A.U.C | Accu. | P. | R. | F1 | A.U.C | Accu. |
| FF++ | No operation | 0.9998 | 0.9996 | 0.9968 | 0.9999 | 0.9968 | 0.9973 | 0.9985 | 0.9979 | 0.9999 | 0.9979 | 0.9982 | 0.9992 | 0.9986 | 0.9999 | 0.9987 | 0.9368 | 0.9424 | 0.9396 | 0.9882 | 0.9395 |
| FF++ | Blur | 0.9807 | 0.9982 | 0.8329 | 0.9831 | 0.7998 | 0.7540 | 0.958 | 0.8438 | 0.9336 | 0.8227 | 0.7151 | 1.0 | 0.8338 | 0.9854 | 0.8007 | **0.7720** | **0.9894** | **0.8673** | **0.9644** | **0.8486** |
| FF++ | Addition of Noise | 0.5 | 0.5 | 0.5 | 0.5 | 0.5 | 0.5 | 0.0095 | 0.0186 | 0.4979 | 0.50 | 0.7489 | 0.1103 | 0.1923 | 0.6546 | 0.5366 | 0.662418 | 0.808201 | 0.727989 | 0.763868 | 0.693863 |
| FF++ | Compression | 0.9376 | 0.0764 | 0.1419 | 0.9300 | 0.5381 | 0.9973 | 0.231 | 0.3751 | 0.9002 | 0.6151 | 0.9992 | 0.6947 | 0.8196 | 0.9914 | 0.8471 | 0.9713 | 0.8651 | 0.9159 | 0.9839 | 0.9206 |
| DFDCPreview | No operation | 0.9999 | 0.9986 | 0.9989 | 0.9999 | 0.999 | 1.0 | 0.9986 | 0.9993 | 1.0 | 0.9993 | 1.0 | 0.9993 | 0.9996 | 0.9999 | 0.9996 | **0.998** | **0.9986** | **0.9983** | **0.9999** | **0.9998** |
| DFDCPreview | Blur | 0.9967 | 0.72 | 0.8372 | 0.9965 | 0.86 | 1.0 | 0.8273 | 0.9055 | 0.9957 | 0.9136 | 1.0 | 0.9433 | 0.9708 | 0.9999 | 0.9716 | 0.976215 | 0.972284 | 0.97425 | 0.982014 | 0.97425 |
| DFDCPreview | Addition of Noise | 0.5598 | 1.0 | 0.6666 | 0.6015 | 0.5 | 0.2727 | 0.01 | 0.019 | 0.4388 | 0.4916 | 0.4978 | 0.99 | 0.6625 | 0.6841 | 0.4956 | **0.9788** | **0.986** | **0.9824** | **0.9985** | **0.982** |
| DFDCPreview | Compression | 0.9943 | 1.0 | 0.7587 | 0.9938 | 0.9939 | 0.9247 | 0.9986 | 0.9602 | 0.9989 | 0.9586 | 0.9621 | 0.9993 | 0.9803 | 0.9998 | 0.98 | 0.9926 | 0.9893 | 0.9909 | 0.9995 | 0.9910 |
| Celeb-DF | No operation | 0.9848 | 0.9376 | 0.9406 | 0.9814 | 0.9408 | 0.8483 | 0.9008 | 0.8738 | 0.9431 | 0.8698 | 0.9235 | 0.9298 | 0.9266 | 0.9823 | 0.9264 | **0.976** | **0.9442** | **0.9603** | **0.9928** | **0.9610** |
| Celeb-Df | Blur | 0.9600 | 0.9421 | 0.8912 | 0.9574 | 0.885 | 0.8191 | 0.8792 | 0.8481 | 0.9189 | 0.8425 | 0.8717 | 0.9345 | 0.9020 | 0.9631 | 0.8985 | **0.9502** | **0.9258** | **0.9379** | **0.9812** | **0.9387** |
| Celeb-Df | Addition of Noise | 0.5 | 0.5 | 0.5 | 0.513 | 0.50 | 0.4766 | 0.09312 | 0.15558 | 0.4947 | 0.4954 | 0.6924 | 0.1513 | 0.2484 | 0.6218 | 0.5421 | **0.6649** | **0.7581** | **0.7084** | **0.7365** | **0.6721** |
| Celeb-Df | Compression | 0.9344 | 0.534 | 0.6881 | 0.9328 | 0.758 | 0.7512 | 0.9086 | 0.8228 | 0.9093 | 0.8044 | 0.9192 | 0.778 | 0.8427 | 0.9481 | 0.8548 | **0.9203** | **0.9252** | **0.9227** | **0.9793** | **0.9226** |

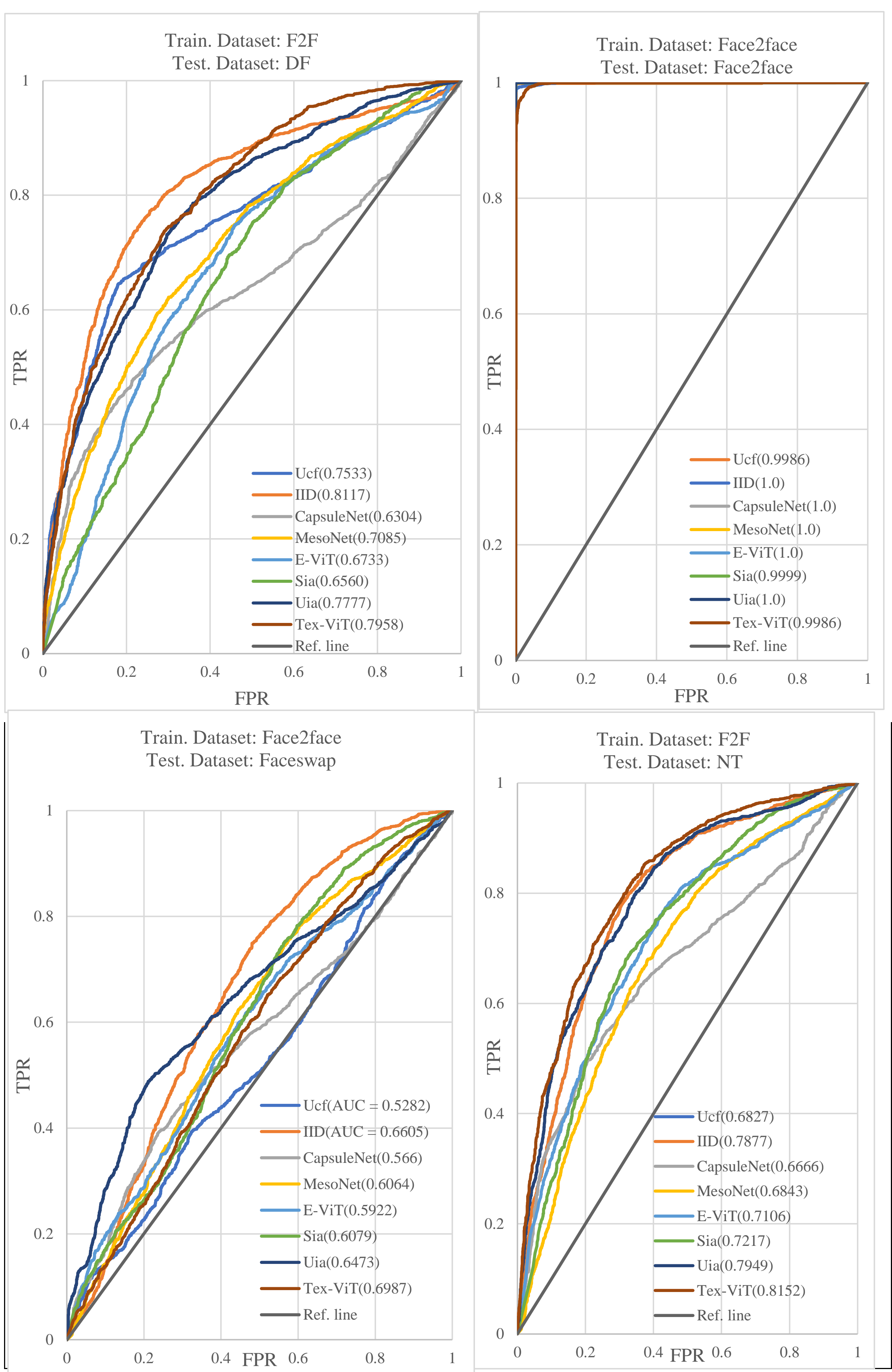

Figure 5: ROC curve generated for the model that was trained using the Face2face dataset and subsequently evaluated across the other categories of FF++. Values shown in the bracket are the AUC values.

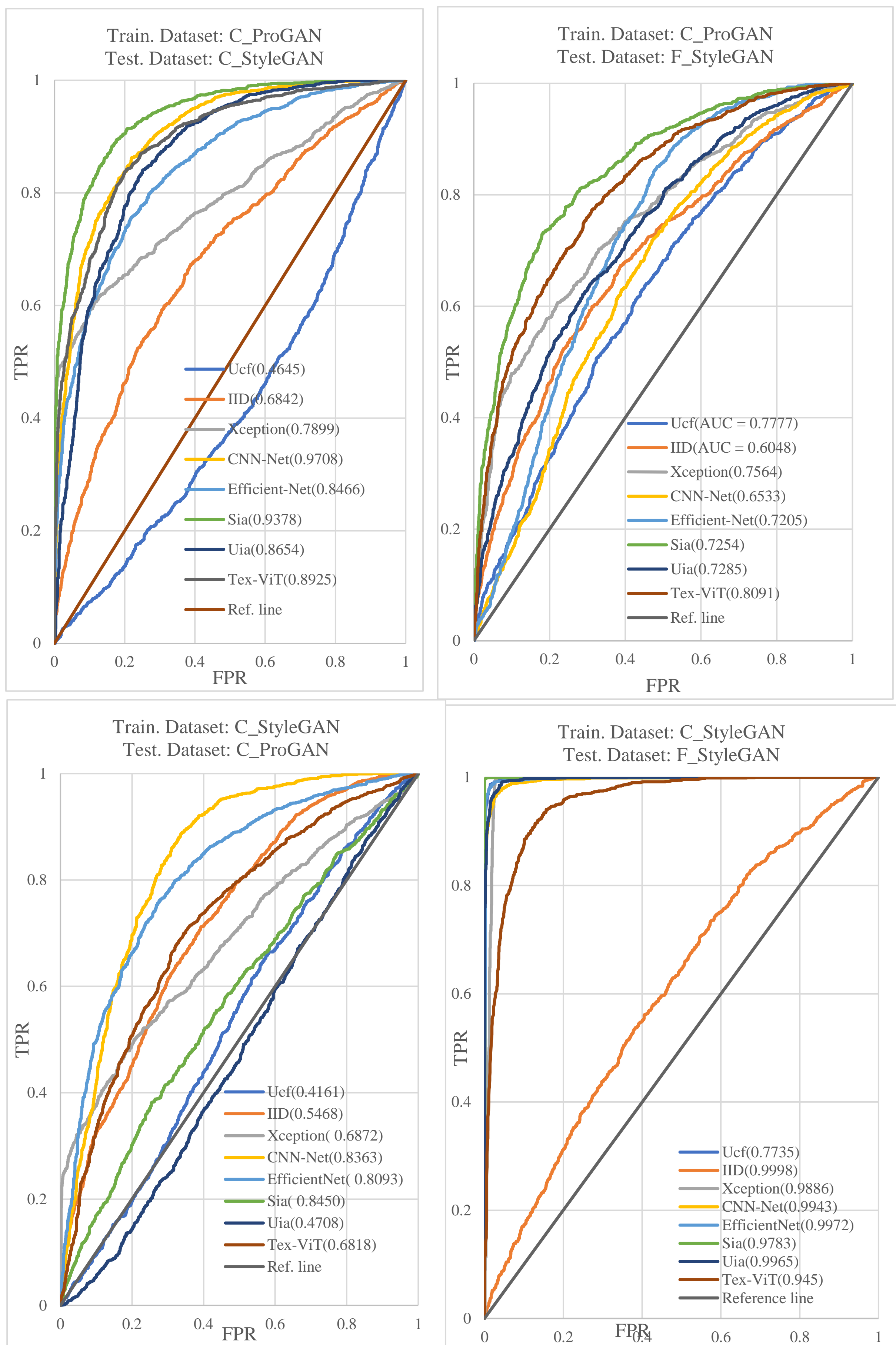

Figure 6: ROC curve analysis for the model trained and evaluated on various types of GAN-generated images. Values shown in the bracket are the AUC values.

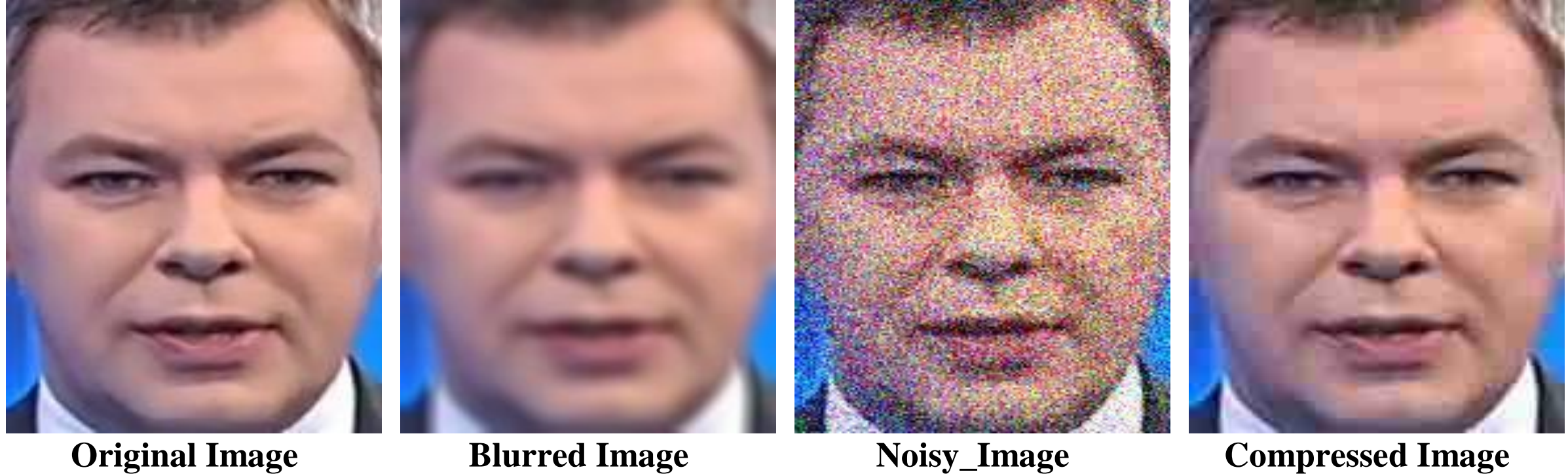

Figure 7: Image undergoes different post-processing operations

## 6 Ablation Studies

The ablation study used to validate each Tex-ViT component is presented in this section. The experiments are conducted using various categories from the Faceforensics++ dataset. The components undergo training across three categories of FF++, while the fourth category serves as the testing ground. This process is executed recursively for each category.

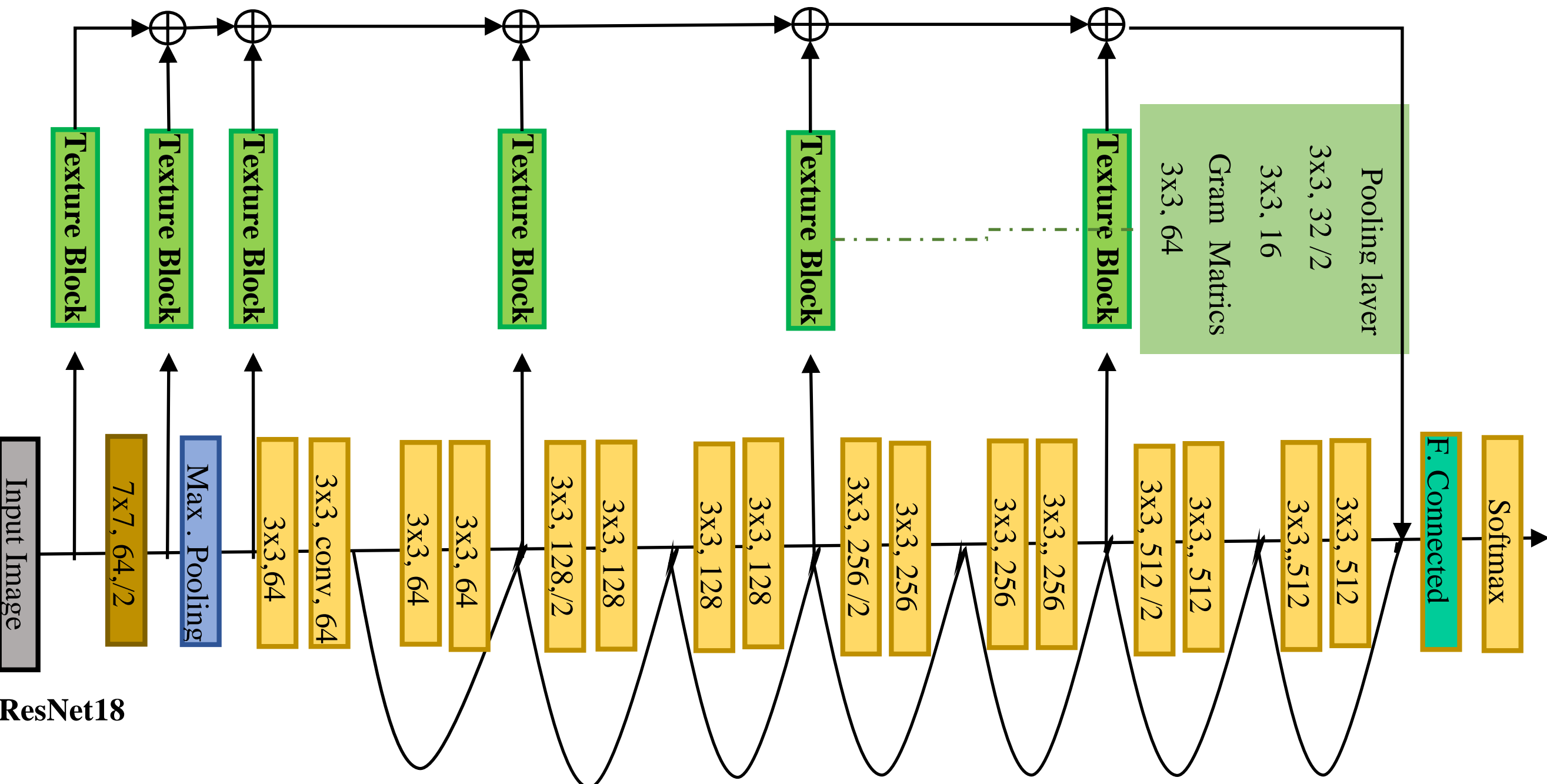

Figure 8: The ResNet18 model design includes a Texture block, in which the characteristics from both branches are combined using concatenation to make the final prediction.

### 6.1 ResNet with Texture module

This section outlines the ResNet18 architecture as the foundational framework, with the texture block computed before each downsampling step. The texture features are also set to be concatenated and eventually integrated into the main branch. The concatenated features are subsequently processed via a sigmoid layer and a fully connected layer to yield the final prediction. Figure 8 illustrates the component in detail. The classification can be found in Table 8. The typical precision performance ranges from 70 to 78 percent, indicating that texture, even without an attention mechanism, is a significant feature for achieving explicit classification. The iterative process of combining and integrating these features with the main branch enables the model to effectively address local and global texture anomalies, thereby improving the overall detection of these features. While CNN-based architectures demonstrate significant effectiveness in executing a wide range of vision tasks, notable generalization limitations have been identified. Their adaptability to a specific context limits their ability to learn global information, making them more susceptible to object scale or layout changes, as they depend heavily on local receptive fields. Their fixed induction bias and lack of capacity for dynamic attention limit adaptability in modeling intricate transformations. However, including a texture branch addresses this limitation of CNN-based architecture, enabling the model to capture subtle invariant features (i.e., texture). Its computation across various scales facilitates the architecture's ability to model the correlation of multi-scale long-range feature dependencies. Also, the component is experiencing significant difficulties in the face-swap category during classification. This issue can be attributed to inconsistencies at the face-swap boundary, which occur over a short range and may not be adequately captured. The model effectively captures the discriminative features across various categories, indicating that the diverse manipulation methods influence texture.

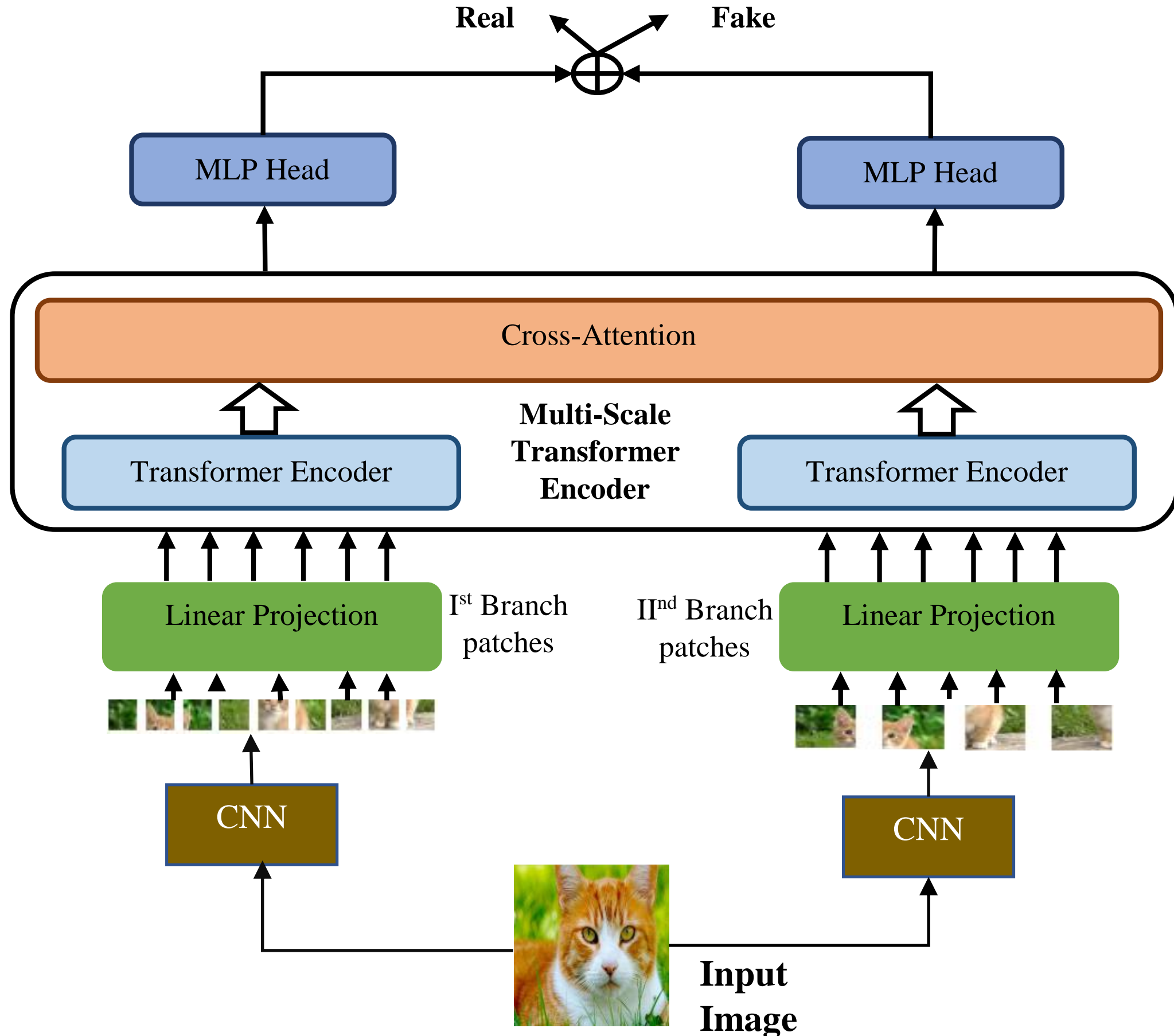

Figure 9: The Dual-branch cross-attention Vision transformer employs a model diagram in which the input image undergoes convolutional neural networks (CNNs) to generate varying-size feature maps.

### 6.2 Dual-branch cross-attention vision transformer

Figure 9 displays an illustration of this component. The input is processed through two convolutional neural networks (CNNs) operating in parallel branches, generating feature maps of varying dimensions. The transformer encoder then receives these feature maps to support a cross-attention mechanism. The final prediction is then generated by concatenating the output of a multi-layer perceptron fed with the features. Table 8 presents the classification results for this component. The accuracy results range from 62% to 70%, indicating performance levels that are suboptimal when compared to the previous component. This suggests that traditional CNN features combined with cross-attention are inadequate for classifying various manipulated images. The overall generalization score of this model falls short compared to that of the previous module; hence, even the two-stream architecture equipped with a cross-attention mechanism and a substantial number of parameters fails to deliver robust generalization capabilities. The model necessitates an external work mechanism designed to enable the modeling of discriminative features, similar to the previously discussed module. Furthermore, the module lacks adaptive mechanisms, which may compromise its effectiveness and versatility across various image adversarial conditions. The score for face swap samples exceeds that of the previous component, demonstrating the substantial influence of the attention mechanism in areas where texture features lack discriminative power. The conclusion regarding this component is that a transformer utilizing cross-attention alone is insufficient to adequately capture the significant latent features inherent across a range of diverse manipulated samples.

### 6.3 Tex-ViT(ResNet+Texture Block + Cross-Attention Transformer)

The proposed framework combines ResNet with a texture block and incorporates a cross-attention transformer. Analysis of components indicates that texture could serve as a significant marker for identifying deepfakes, showing reliability across different manipulation methods. A cross-attention vision transformer, devoid of inductive biases, simultaneously highlights the details of the manipulation. The computation of texture is integrated with conventional CNN features. The features undergo a refinement process, focused attention, and expansion through the cross-attention mechanism, resulting in an effective feature representation and improved performance. Additional experimentation demonstrates that integrating texture features across various scales enhances the model's robustness to different types of image manipulation. The continuous integration and combination of these features with the main branch allow the model to identify local and global texture irregularities proficiently, thus enhancing overall detection precision. Furthermore, incorporating texture features across multiple scales strengthens the model's resilience to image alterations. The data presented in the table indicates that this effective combination achieves an accuracy rate of eighty percent for Deepfake and Face2face samples, exceeds seventy percent for NT, and surpasses sixty percent for face-swap. This document verifies that the model can obtain distinguishing characteristics consistent across various sample types.

Table 8: The classification results for the various components are displayed. Components are recursively tested on the fourth category after being taught on other FF++ categories, where Case A = F2F+FS+NT, Case B = DF+FS+NT, Case C = DF+F2F+NT, Case D = DF+F2F+FS

| Train | Test | Precision | Recall | F1 Score | AUC | Acc |
|---|---|---|---|---|---|---|
| ResNet with Texture block | | | | | | |
| Case A | DF | 0.83 | 0.7045 | 0.7624 | 0.8784 | 0.78 |
| Case B | F2F | 0.882 | 0.7625 | 0.8179 | 0.9207 | 0.83 |
| Case C | FS | 0.4646 | 0.151 | 0.2279 | 0.5686 | 0.48 |
| Case D | NT | 0.7958 | 0.569 | 0.6635 | 0.8091 | 0.71 |
| Cross-Attention Transformer(dual branch). | | | | | | |
| Case A | DF | 0.6578 | 0.7748 | 0.7115 | 0.7759 | 0.7001 |
| Case B | F2F | 0.6784 | 0.6956 | 0.6869 | 0.7489 | 0.6845 |
| Case C | FS | 0.6512 | 0.5512 | 0.5974 | 0.6748 | 0.6306 |
| Case D | NT | 0.7002 | 0.6785 | 0.68917 | 0.7458 | 0.6978 |
| Tex-ViT (Our final model) | | | | | | |
| Case A | DF | 0.7901 | 0.949 | 0.8622 | 0.9275 | 0.8485 |
| Case B | F2F | 0.8364 | 0.8565 | 0.8463 | 0.9169 | 0.8463 |
| Case C | FS | 0.5773 | 0.543 | 0.5596 | 0.5900 | 0.57275 |
| Case D | NT | 0.7739 | 0.777 | 0.754 | 0.8600 | 0.775 |

Table 9: Tex-ViT's computational complexity is compared to well-known computer vision models. The column labeled "parameter" represents the number of trainable parameters, Accuracy score, GPU and CPU time taken for a batch size.

| Model | Accuracy | Parameter (millions) | CPU time(sec) | GPU time(sec) |
|---|---|---|---|---|
| CNN-Net | 52.47 | 23.51 | 1.99s | 0.379s |
| ResNext | 79.55 | 81.41 | 4.61s | 0.338s |
| CapsuleNet | 60.56 | 1.5 | 2.49s | 0.154 |
| EfficientNetV2 | 80.5 | 118.51 | 3.93s | 0.40s |
| XceptionNet | 71.6 | 20.81 | 2.45s | 0.66s |

| MesoInception | 62.3 | 0.028 | 2.1s | 0.72s |
|---|---|---|---|---|
| Swin Transformer | 75.4 | 59.96 | 4.92s | 0.41s |
| ResNet152 | 79.3 | 58 | 3.03s | 0.309s |
| ConvNext | 83.39 | 88.57 | 3.62 | 0.37s |
| Efficient_b7 | 75.49 | 66.34 | 4.07s | 0.35s |
| **Tex-ViT** | **84.85** | **43** | **3.06s** | **1.02s** |

## 7 Computational Complexity Analysis of the Proposed Model

Compared to well-known computer vision models, the computational complexity of the Tex-ViT architecture is examined in this section. The number of trainable parameters, the precision level attained on the DF (FF++) dataset, and the inference times measured on both CPU and GPU are key computational factors that has been taken into account. A tabular depiction of Tex-ViT's complexity study concerning several common computer vision models is shown in Table 9. Tex-ViT is a lightweight model compared to various computer vision models, including EfficientNetv1, and it has demonstrated superior performance across diverse scenarios(Figure 10). The number of trainable parameters in Efficientv2 and ConvNext is relatively high, resulting in an accuracy score of approximately 80%. A greater number of parameters typically enables the model to acquire more discriminative information regarding the manipulation. Deepfake detection models such as MesoInceptionNet, Capsule Net, and CNN-Net are characterised by their lightweight architecture; however, their performance is compromised in varied scenarios. The execution time for a batch size of 32 is seconds during the inference phase. The increased number of parameters in heavy models necessitates a longer execution time than in light-weight models. Xception is a lightweight model comprising 20 million parameters, achieving an accuracy score of 70%.

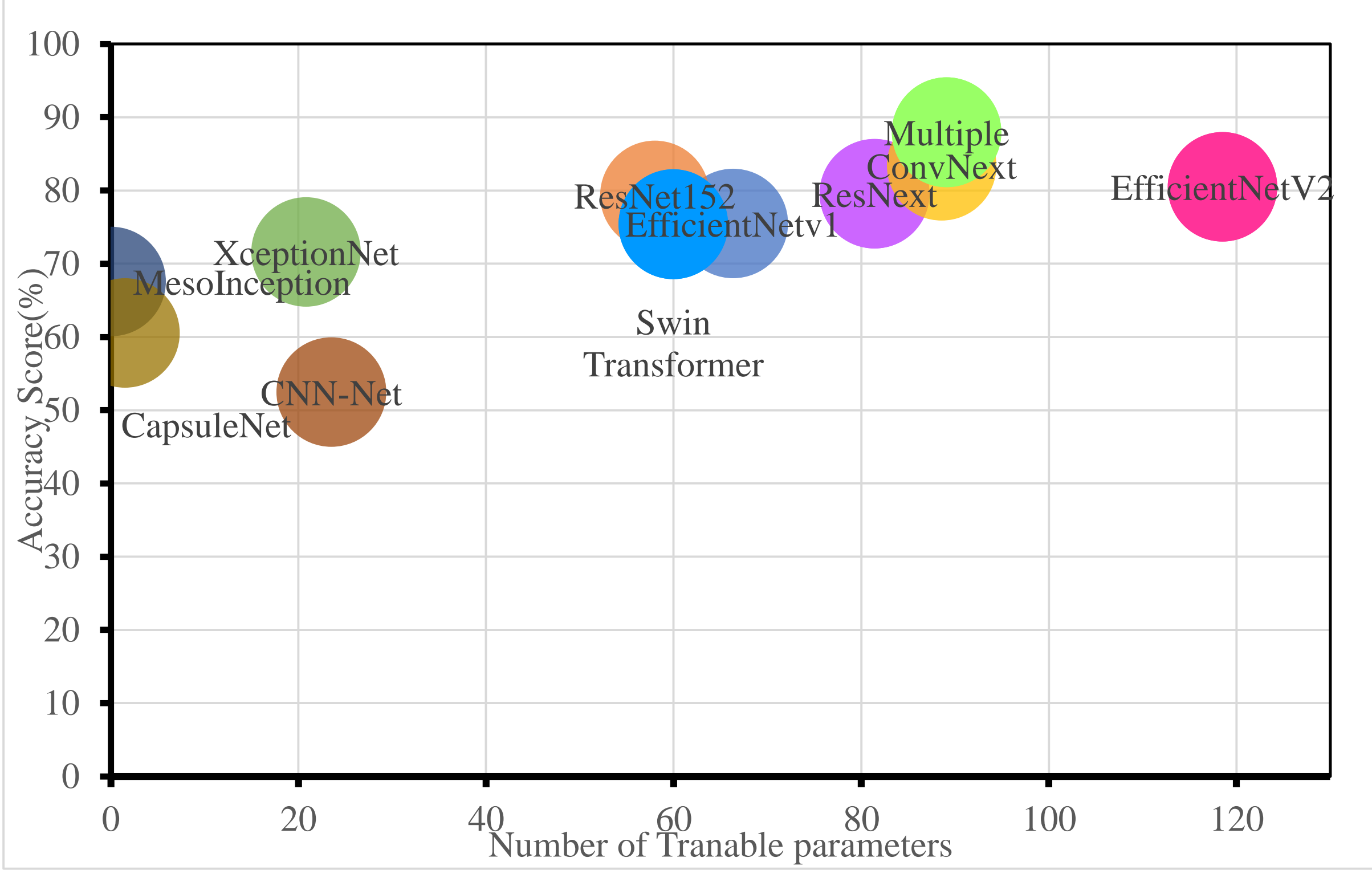

Figure 10: Tex-ViT's complexity analysis compared to well-known computer vision models.

## 8 Limitation and Future work

The approach demonstrates limitations in its effectiveness in cross-domain scenarios, particularly when assessing images from the faceswap category. The model demonstrates limited performance regarding the noise samples, indicating a potential area for future enhancement. Future efforts will concentrate on improving the model's precision in the FS manipulation domain of FF++. This could involve the incorporation of additional modules to enhance feature learning. Additionally, enhancing the model's robustness concerning compressed data samples will be a focus for future work.

## 9 Conclusion

This paper presents an empirical experimentation from various perspectives to state that fake images has smoother texture characteristics rather than real images. Additionally, it has been observed that texture correlation is not maintained in instances of synthetic images, resulting in a tendency for these images to exhibit smoother surfaces. The proposed Tex-ViT integrates conventional CNN features derived from ResNet and texture modules utilizing ResNet features. The output from these two parallel branches functions as an input for the dual-branch vision transformers, which operate on patches utilizing the cross-attention mechanism. The experimental results indicate that the model demonstrates good performance in the cross-domain context of FF++ datasets and various GAN image datasets, surpassing other leading models in the field. The experimentation conducted on various post-processing image scenarios indicates the model's robustness against adversarial attacks. The experiments demonstrate that the model identifies and retains the common discriminative features across multiple fake images. The evaluation indicates that texture is an invariant feature consistent across different manipulation methods. Learning this feature is expected to lead to improved performance for the model. The model exhibits a low score in the FS category of FF++. Experimentation is conducted on various types of GAN image datasets, demonstrating superior performance compared to other state-of-the-art models. The model demonstrates enhanced learning capabilities across various feature spaces. Various post-processing image scenarios were subjected to experimentation, revealing that the model demonstrates sufficient robustness against different adversarial operations. Nevertheless, the model requires enhancement in its performance for compressed scenarios; however, its score remains superior to that of other state-of-the-art models. The experiments demonstrate that the model identifies and retains the common discriminative features across multiple fake images.

## 10 Author’s contributions, compliance with Ethical standards, declaration of competing interests, availability, and informed consent of data.

*Author’s contribution*

**Deepak Dagar:** Software, Validation, Investigation, Data Curation, Writing – Original Draft, Visualization.

**Dinesh Kumar Vishwakarma:** Conceptualization, Methodology, Formal Analysis, Resources, Writing – Review & Editing, Supervision, Project Administration, Funding Acquisition.

*Compliance with Ethical Standards*

The authors affirm that the manuscript has adhered to the ethical standards of the Journal of Information Security and Applications.

*Competing interests*

The authors affirm that they do not possess any identifiable conflicting financial or non-financial interests or personal ties that could have potentially influenced the findings presented in the research.

*Research Data Policy and Data Availability Statements*

Given that the dataset and its corresponding code are open-source and accessible to the public. Therefore, it is unnecessary to acquire informed consent from the pertinent stakeholders before use the dataset.

*Funding Declaration*
There was no funding taken for this research work.